\documentclass[10pt, oneside, leqno]{article}
\usepackage[left=35mm, right=35mm, top=35mm, bottom=40mm]{geometry}
\usepackage{graphicx}
\usepackage{amsmath, amssymb, amsfonts, mathtools}
\usepackage[square, authoryear]{natbib}
\usepackage{enumitem}
\usepackage{authblk}
\usepackage{titlesec}
\usepackage[labelsep=period,labelfont=bf,font=small]{caption}
\usepackage{bm}
\usepackage[hidelinks]{hyperref}

\author{Christian W\"ulker\thanks{e-mail: \texttt{christian.wuelker@jhu.edu}} }
\author{Gregory S.\ Chirikjian}
\affil{\small{Department of Mechanical Engineering\\Johns Hopkins University, Baltimore, MD, USA}}

\setlist[enumerate]{  
  wide=\parindent
}

\makeatletter
\def\moverlay{\mathpalette\mov@rlay}
\def\mov@rlay#1#2{\leavevmode\vtop{%
   \baselineskip\z@skip \lineskiplimit-\maxdimen
   \ialign{\hfil$\m@th#1##$\hfil\cr#2\crcr}}}
\newcommand{\charfusion}[3][\mathord]{
    #1{\ifx#1\mathop\vphantom{#2}\fi
        \mathpalette\mov@rlay{#2\cr#3}
      }
    \ifx#1\mathop\expandafter\displaylimits\fi}
\makeatother

\title{\Large\textbf{Quantizing Euclidean motions via double-coset decomposition}}

\date{\small\today}

\begin{document}

\maketitle

\begin{abstract}
Concepts from mathematical crystallography and group theory are used here to quantize the group of rigid-body motions, resulting in a ``motion alphabet'' with which to express robot motion primitives. From these primitives it is possible to develop a dictionary of physical actions. Equipped with an alphabet of the sort developed here, intelligent actions of robots in the world can be approximated with finite sequences of characters, thereby forming the foundation of a language in which to articulate robot motion. In particular, we use the discrete handedness-preserving symmetries of macromolecular crystals (known in mathematical crystallography as Sohncke space groups) to form a coarse discretization of the space $\textnormal{SE}(3)$ of rigid-body motions. This discretization is made finer by subdividing using the concept of double-coset decomposition. More specifically, a very efficient, equi\-volumetric quantization of spatial motion can be defined using the group-theoretic concept of a double-coset decomposition of the form $\Gamma \backslash \textnormal{SE}(3) / \Delta$, where $\Gamma$ is a Sohncke space group and $\Delta$ is a finite group of rotational symmetries such as those of the icosahedron. The resulting discrete alphabet is based on a very uniform sampling of $\textnormal{SE}(3)$ and is a tool for describing the continuous trajectories of robots and humans. The general ``signals to symbols'' problem in artificial intelligence is cast in this framework for robots moving continuously in the world, and we present a 
coarse-to-fine search scheme here to efficiently solve this 
decoding problem in practice. 

\end{abstract}

\section{Introduction}
\label{sec:introduction}
The aim of this paper is to develop a ``motion alphabet'' with which to express robot motion primitives. From these primitives one can develop a dictionary of physical actions. Two main themes from the theory of Lie groups are used to construct this alphabet (or quantization) and to efficiently solve the ``signals to symbols'' problem in this context: (1) the decomposition of a Lie group into cosets, double cosets, and corresponding fundamental domains; (2) the possibility to construct such fundamental domains as Voronoi or Voronoi-like cells.

Equipped with an alphabet of the sort developed in this paper and the associated algorithms for efficiently rounding off continuous motions to nearby discrete representatives, intelligent actions of robots in the world can be approximated with finite sequences of characters, thereby forming the foundation of a language in which to articulate robot motion.

At the macroscopic scale, the world can be thought of as continuous. Coarse descriptions of this continuous world are used by intelligent systems (\textit{e.\,g.}, humans and computers) to classify objects, actions, and scenarios. The classical ``signals to symbols'' problem in artificial intelligence (AI) seeks to bin everything in the continuous world into countable classes characterized by strings of discrete symbols, such as the letters in an alphabet. In a sense, this is the inverse problem of what the genetic code does, since the finite alphabet $\{\textnormal{A},\textnormal{C},\textnormal{G},\textnormal{T}\}$ encodes the morphology and metabolism of every living creature that moves in the continuous world and processes information with an analog brain.

Discrete alphabets (including the Roman alphabet, the radicals that form characters in Asian languages, sign language, etc.) form the basis for all human language. All discrete characters can be reduced to a binary code, \textit{e.\,g.}, ASCII or Morse code in the case of western letters. The efficiency of an alphabet (or a code) depends on how much information can be conveyed with a given number of symbols, and how difficult it is to convey those symbols. For example, since the letter `E' is the most often used symbol in the English language, it is represented by a singe `$\cdot$' in Morse code.

As an example of how discrete symbols classify the world, consider how each of the following sentences describes a class of situations in which there is continuous freedom which becomes more restricted as the number of discrete symbols increases:

\begin{enumerate}
\item The cat is on the table.
\item The black cat is sitting on the table.
\item The black cat is sitting on the table and looking at you.
\end{enumerate}

In 1, the cat could be of any color and lying down, sitting, or standing in an infinite variety of postures. The color and posture of the cat gets somewhat restricted as the sentences get longer, but we still do not know how the cat is sitting, what its tail is doing, how big it is, its weight, the color of its eyes, etc. Indeed, the sentence would need to be quite long to hone in on what is actually going on, hence the old saying ``One picture is worth a thousand words.''

The basic problem that must be reconciled by intelligent robots is to approximate, or round off, continuous objects and actions within a discrete descriptive framework, and then truncate the discrete description at some finite level given limitations on computational and sensing resources. This is true both in machine-learning (ML) frameworks such as deep learning and classical AI. A specific kind of rounding off (of Euclidean motions) is the main subject of this paper, which is about making precise the round-off statement 
\begin{equation*}
\textnormal{discrete description} \,=\, [\textnormal{continuous description}]
\end{equation*}
in the context of motions of objects and intelligent agents in the world.

The remainder of this paper is structured as follows: In Section \ref{sec:literature_review}, a review of the immense literature on machine intelligence as it applies to intelligent robot action in the world is provided. This is a rapidly changing field and impossible to capture in full detail, but some classical highlights are covered. Section \ref{sec:group_theory} reviews some relevant aspects of abstract group theory. This is made concrete in Section \ref{sec:rigid_body_motions} which focuses on the group of rigid-body motions. Section \ref{sec:crystallographic_groups} reviews crystallographic symmetry, which is a source of discrete symbols from which to construct a motion alphabet. In Section \ref{sec:motion_alphabet}, 
we develop a motion alphabet by ``dividing up'' the group $\textnormal{SE}(3)$ of rigid-body motions via a fine double-coset decomposition based on a crystallographic Sohncke space group and the subgroup of rotational symmetries of the icosahedron. Section \ref{sec:decoding_algorithms} presents other choices for motion alphabets and solves the decoding problem efficiently by introducing a 
coarse-to-fine search scheme.

\section{Literature review}
\label{sec:literature_review}
The recognition of human (and humanoid-robot) actions has been studied from many different perspectives including \citep{Yilmaz05,active-vision-last,Turaga,Liu,skeletons,Hausman}. Probabilistic graphical models \citep{Koller}, generative models \citep{Leyton}, and recently ``SE3-nets'' \citep{se3-nets} have been used to describe motion uncertainty in the context of learning. Work in vision and reasoning use concepts of quotient operations to mod out irrelevant information \citep{Soatto, Li10, quotient-space-reasoning}. Group-theoretic methods (and abstract algebra more generally) can be found sprinkled throughout the AI literature \citep{Hartmanis,ginzburg,Arbib,perceptrons,yanxi1,Henderson09,domingos}.

Of particular importance in the current context is the relationship between artificial intelligence and machine learning. AI arose as a branch of cybernetics, focusing on artificial aspects of reasoning and cognition, thereby leading to a redefinition of cybernetics to focus on information and control. Machine learning (and particularly deep neural networks) led by Hinton, LeCun, Bengio, and others can be viewed as an alternative to classical AI \citep{LeCun98,Hinton1,Bengio,Bengio09,LeCun14}. Recently, geometric and algebraic methods are being explored in some forms of machine learning \citep{Vejdemo-Johansson,icml2,icml3}.

The goal of this paper is to develop an alphabet of basic motions from which to construct discrete words that capture the essence of a continuous motion/action. These actions will form part of a dictionary. This dictionary will, in the future, serve as the knowledge base for an expert system that will enable the robot to function at first use ``right out of the box.''

\section{Some relevant aspects of group theory}
\label{sec:group_theory}
The alphabets constructed in this paper consist of carefully chosen elements of the group of rigid-body motions, drawn from fundamental domains of double-coset spaces. Since this terminology might not be familiar to some readers with an interest in the topic, the relevant concepts from group theory are reviewed here. The concept of a group itself is assumed to be known. 

\subsection{Definitions and properties}

Let a group $G$ and a subgroup $H < G$ be given. For an element $g \in G$, the corresponding \emph{left coset} is defined as
\begin{equation*}
gH \,\coloneqq\, \{g h \,:\, h \in H\}.
\end{equation*}
Analogously, the respective \emph{right coset} is defined as
\begin{equation*}
Hg \,\coloneqq\, \{h g \,:\, h \in H\}.
\end{equation*}
A group can be divided into cosets (left or right), all with the same number of elements, and the set of all these cosets is called a \emph{coset space}. In general, the left-coset space
\begin{equation*}
G / H \,\coloneqq\, \{gH \,:\, g \in G\}
\end{equation*}
and the right-coset space
\begin{equation*}
H \backslash G \,\coloneqq\, \{Hg \,:\, g \in G\}
\end{equation*}
are different from each other, except for when $H$ is normal in G ($H \triangleleft G$), in which case $gH = Hg$ for all $g \in G$. In this special case $G / H = H \backslash G \eqqcolon \frac{H}{G}$ is a \emph{factor} (or \emph{quotient}) \emph{group}. 

Given two subgroups $H,K < G$, it is also possible to define \emph{double cosets}
\begin{equation*}
HgK \,\coloneqq\, \{h g k \,:\, h \in H,\, k \in K\}, \quad g \in G,
\end{equation*}
and the corresponding \emph{double-coset space},
\begin{equation*}
H \backslash G / K \,\coloneqq\, \{HgK \,:\,g \in G\}.
\end{equation*}

Given a left-coset decomposition with respect to a subgroup $H$, it is possible to define a (non-unique) \emph{fundamental domain}
\begin{equation*}
F_{G / H} \,\subset\, G
\end{equation*}
consisting of exactly one element per left coset. 
Since a group can be partitioned into disjoint cosets, it is
\begin{equation*}
G \,=\, \bigcup_{g \in F_{G/H}} gH
\end{equation*}
and
\begin{equation}\label{eq:right_coset_fundamental_decomposition}
G \,=\, \bigcup_{h \in H} F_{G / H} h
\end{equation}
(and analogously in the right-coset case). It is important to note that we are generally dealing with unions of disjoint sets in this paper. When $H \triangleleft G$, such fundamental domain is a group with respect to the original group operation modulo $H$, and this group is isomorphic to $G / H$.

Given two subgroups $H, K < G$, the corresponding double-coset decomposition is
\begin{equation*}
G \,=\, \bigcup_{g \in F_{H \backslash G / K}} HgK
\end{equation*}
and we further have that
\begin{equation}\label{eq:double_coset_fundamental_decomposition}
G \,=\, \bigcup_{h \in H} \bigcup_{k \in K} h F_{H \backslash G / K} k,
\end{equation}
where $F_{H \backslash G / K}$ is a fundamental domain for the double-coset space $H \backslash G / K$ consisting of exactly one element per double coset. When $G$ is a \emph{Lie group} (\textit{i.\,e.}, when $G$ has the structure of a differentiable manifold, and when further the group operation and inversion are compatible with this smooth structure), and $H,K$ are \emph{discrete} subgroups, then such fundamental domains $F_{G / H}$ and $F_{H \backslash G / K}$ will have the same dimensionality as $G$, but smaller volume.

When $G$ is a Lie group, one way to construct fundamental domains is as Voronoi-like cells: 
Since $G$ is a smooth manifold, a proper distance function (metric) $\rho$ exists, 
and we can define
\begin{equation}\label{eq:pre_Voronoi_coset}
\begin{aligned}
F_{G / H}^\circ \,&\coloneqq\, \{g \in G \,:\, \rho(e,g) < \rho(e,gh) ~\,\textnormal{for all}\,~ h \in H \setminus \{e\}\},\\
F_{H \backslash G}^\circ \,&\coloneqq\, \{g \in G \,:\, \rho(e,g) < \rho(e,hg) ~\,\textnormal{for all}\,~ h \in H \setminus \{e\}\},
\end{aligned}
\end{equation}
where $e$ is the identity element of $G$, and when $H \cap K = \{e\}$,
\begin{equation}\label{eq:pre_Voronoi_double_coset}
F_{H \backslash G / K}^\circ \,\coloneqq\, \{g \in G \,:\, \rho(e,g) < \rho(e,hgk) ~\,\textnormal{for all}\,~ (h,k) \in H \times K \setminus \{(e,e)\}\}.
\end{equation}
Above we have defined the \emph{interior} of fundamental domains. The union of the corresponding shifts of these open sets will in general not completely exhaust G (cf.\ \eqref{eq:right_coset_fundamental_decomposition} and \eqref{eq:double_coset_fundamental_decomposition}). However, a set of measure zero will at most be missing, which is not relevant for our practical purposes.


Group theory has been a cornerstone in all areas of the physical sciences including crystallography, classical mechanics, quantum mechanics, chemistry, and particle physics. 
Moreover, in classical works on finite automata theory, attempts were made to incorporate the theory of finite groups \citep{Hartmanis,ginzburg,Arbib}. Many roboticists and computer vision researchers know about the special Euclidean group $\textnormal{SE}(3)$, which is a Lie group that describes rigid-body motions. This will be discussed later, after first reviewing the group of pure rotations.

\subsection{A concrete example: the rotation group}\label{subsec:SO(3)}

In the following, the abstract definitions are illustrated with a concrete example. The set of $3 \times 3$ rotation matrices is called the \emph{special orthogonal group}, and is denoted as $\textnormal{SO}(3)$. That is, 
\begin{equation}\label{eq:SO(3)}
\textnormal{SO}(3) \,\coloneqq\, \{R \in \mathbb{R}^{3\times 3} \,:\, R^\textnormal{T} R = \mathbb{I} ~\,\textnormal{and}~ \det R = 1\},
\end{equation}
where $\mathbb{I}$ is the identity matrix. Here the group operation is simply matrix multiplication. It is not difficult to show that 
$\mathbb{I}$ is the group identity, given any two $R_1, R_2 \in \textnormal{SO}(3)$ that the matrix product $R_1 R_2 \in \textnormal{SO}(3)$, and that $R^{-1} = R^\textnormal{T}$ is the inverse of $R \in \textnormal{SO}(3)$.

Explicitly for $\textnormal{SO}(3)$, elements of the associated Lie algebra $\mathfrak{so}(3)$, which correspond to infinitesimal rotations, are \emph{skew-symmetric matrices}
\begin{equation*}
X
\,=\,
\begin{bmatrix}
\begin{array}{ccc}
0 & -x_3 & x_2 \\
x_3 & 0 & -x_1 \\
-x_2 & x_1 & 0
\end{array}
\end{bmatrix}\!,
\end{equation*}
and the \emph{matrix exponential} (or \emph{exponential map}) gives
\begin{equation*}
R(\boldsymbol{x}) \,=\, \exp(X) \,=\, \mathbb{I} + \frac{\sin \|\boldsymbol{x}\|_2}{\|\boldsymbol{x}\|_2} X + \frac{1 - \cos \|\boldsymbol{x}\|_2}{\|\boldsymbol{x}\|_2^2} X^2,
\end{equation*}
where $\boldsymbol{x} = (x_1, x_2, x_3)^\textnormal{T} \eqqcolon X^\vee$ is the dual vector corresponding to $X$. The opposite operation gives $\widehat{\boldsymbol{x}} \coloneqq X$.
The parameters $x_1$, $x_2$, and $x_3$ can be thought of as Cartesian coordinates in the Lie algebra $\mathfrak{so}(3)$, and when these coordinates are restricted to the range $\|\boldsymbol{x}\|_2 \leq \pi$ they can be used to parameterize all of $\textnormal{SO}(3)$ through the exponential map. When $\|\boldsymbol{x}\|_2 = \pi$ the point is at the boundary. In such a case $\boldsymbol{x}$ and $-\boldsymbol{x}$ describe the same rotation, and so one model for $\textnormal{SO}(3)$ is that of a solid ball of radius $\pi$ in Euclidean space, with antipodal points identified as being equivalent, or ``glued.''

The inverse map for each is the \emph{matrix logarithm}. This degenerates when the rotation angle $\theta \coloneqq \|\boldsymbol{x}\|_2$ is $\pi$. By restricting the discussion to the case when $\theta < \pi$, the logarithm is uniquely defined on a subset of $\textnormal{SO}(3)$ depleted by the set of measure zero defined by $\theta = \pi$. This depletion will have no effect on our formulation. Indeed, we can define the metric
\begin{equation*}
\rho(R_1, R_2) \,\coloneqq\, \|\log^\vee (R_1^\textnormal{T} R_2)\|_2
\end{equation*}
when $R_1^\textnormal{T} R_2$ is not a rotation by $\pi$, and otherwise $\rho(R_1, R_2) \coloneqq \pi$.

\begin{figure}[t]
\centering
\includegraphics[width=0.32\textwidth]{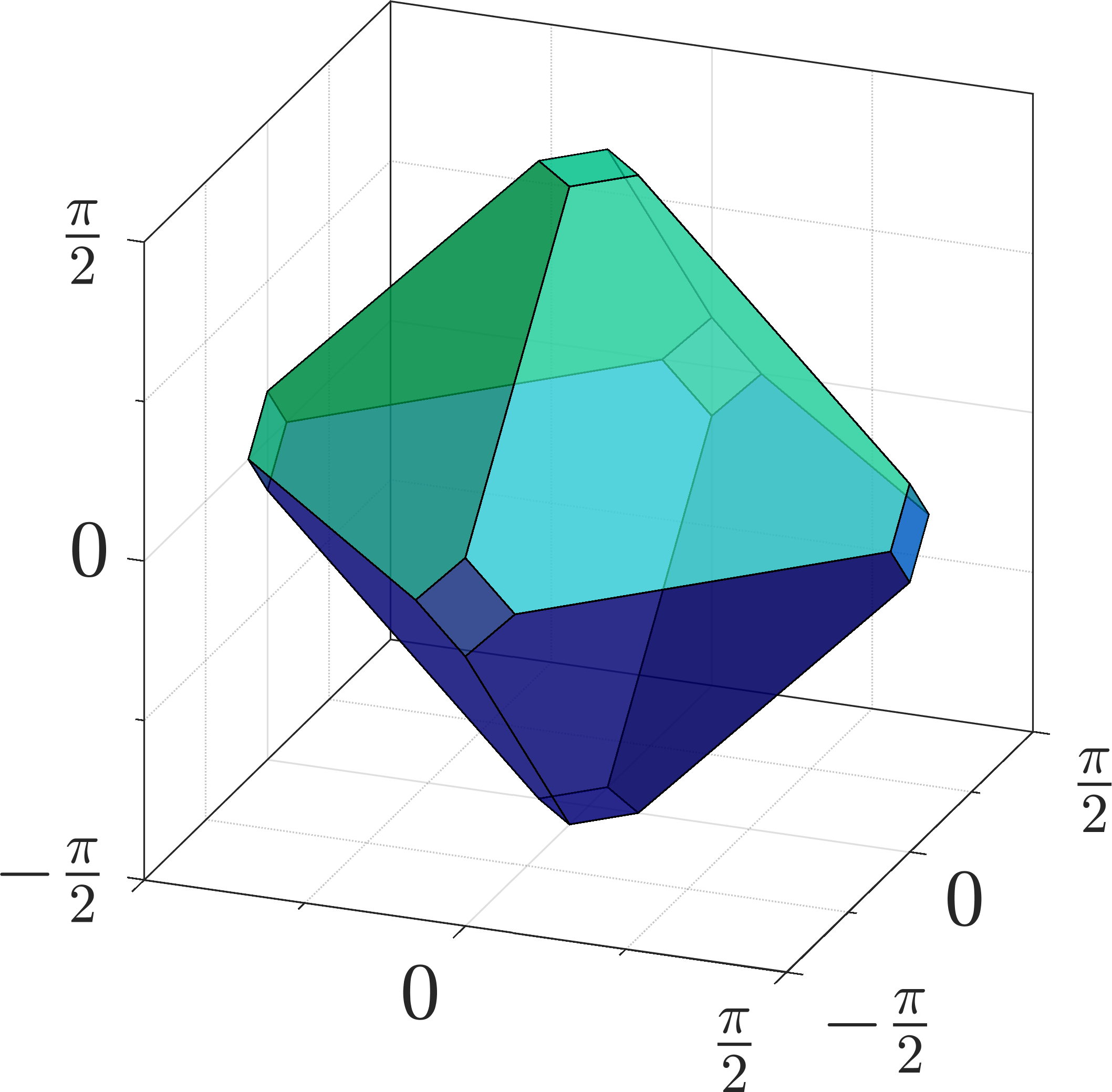}\hfill
\includegraphics[width=0.32\textwidth]{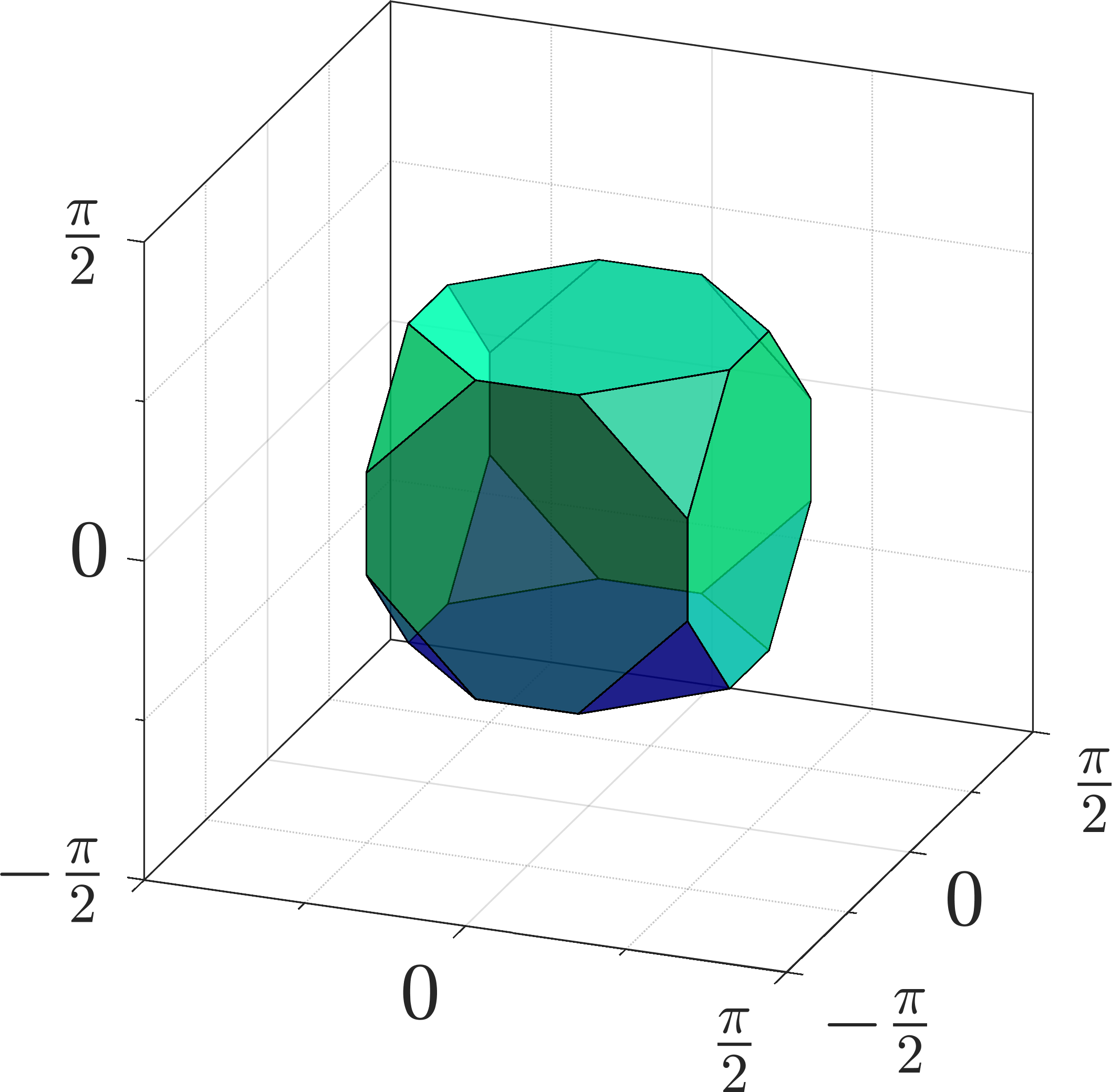}\hfill
\includegraphics[width=0.32\textwidth]{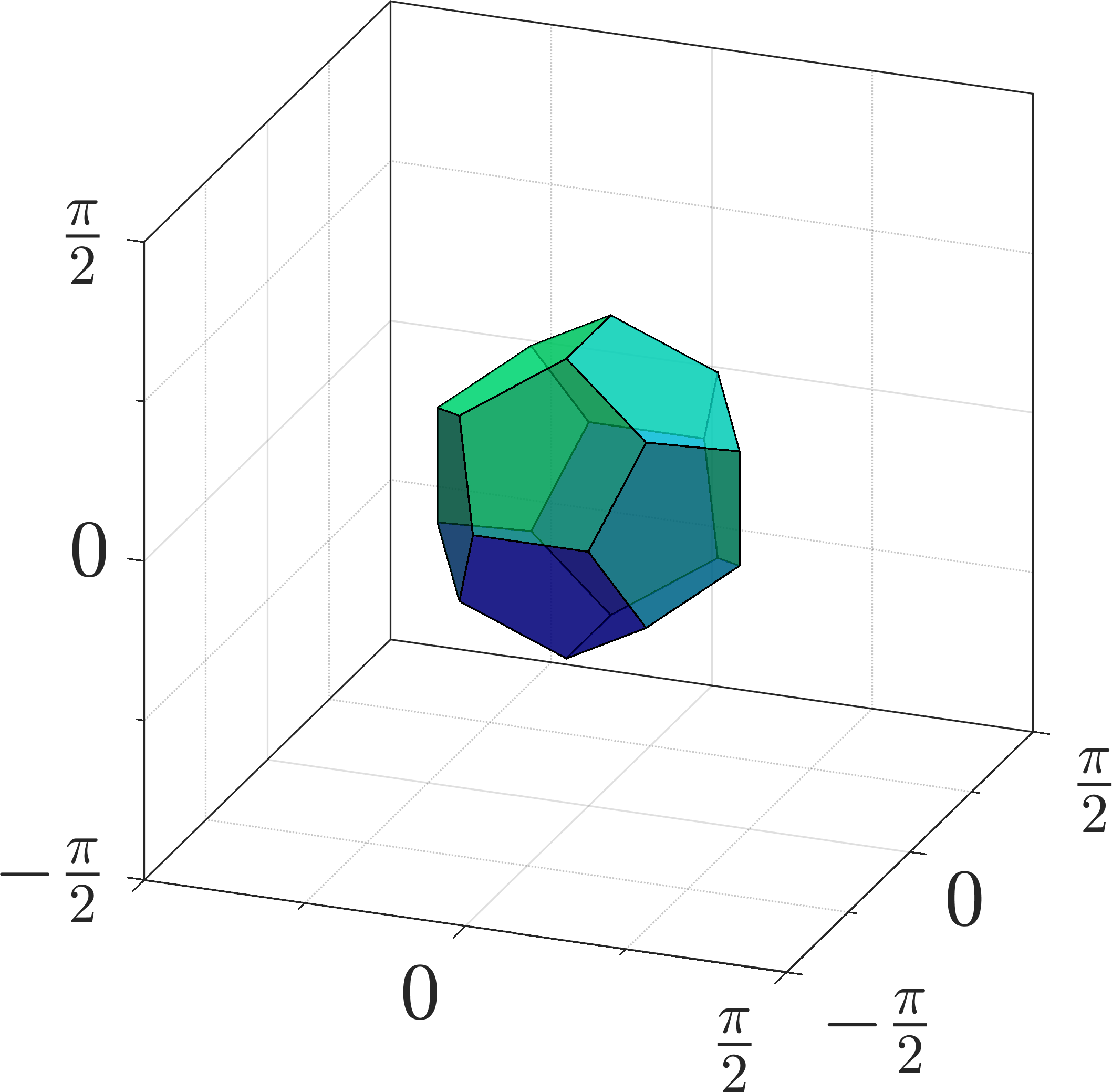}
\caption{Fundamental domain $F_{H \backslash \textnormal{SO}(3)}$ with $H$ as the group of (from left to right) tetrahedral, octahedral, and icosahedral symmetries, constructed as Voronoi cells, viewed in exponential coordinates.}
\label{fig:tiles}
\end{figure}

It is interesting to note that 
the above distance function $\rho$ for $\textnormal{SO}(3)$ is \emph{bi-invariant}, \textit{i.\,e.},
\begin{equation*}
\rho(R_1, R_2) \,=\, \rho(R R_1, R R_2) \,=\, \rho(R_1 R, R_2 R), \quad\quad R \in \textnormal{SO}(3).
\end{equation*}
Using this particular metric $\rho$ it is possible to construct \emph{Voronoi cells} (in the classical sense) in $\textnormal{SO}(3)$ for 
fundamental domains $F_{H \backslash \textnormal{SO}(3)}$ and $F_{H \backslash \textnormal{SO}(3) / K}$, because then \eqref{eq:pre_Voronoi_coset} and \eqref{eq:pre_Voronoi_double_coset} become
\begin{equation}\label{eq:SO(3)_coset_Voronoi}
F_{\textnormal{SO}(3) / H}^\circ \,=\, F_{H \backslash \textnormal{SO}(3)}^\circ \,=\, \{R \in \textnormal{SO}(3) \,:\, \rho(R,\mathbb{I}) < \rho(R,h) ~\,\textnormal{for all}\,~ h \in H \setminus \{\mathbb{I}\}\}\\
\end{equation}
and
\begin{equation}\label{eq:SO(3)_double_coset_Voronoi}
F_{H \backslash \textnormal{SO}(3) / K}^\circ \,=\, \{R \in \textnormal{SO}(3) \,:\, \rho(R,\mathbb{I}) < \rho(R,hk) ~\,\textnormal{for all}\,~ (h,k) \in H \times K \setminus \{(\mathbb{I},\mathbb{I})\}\},
\end{equation}
respectively.

Of particular interest to us are the cases where $H$ is one of the finite groups of rotational symmetries of the Platonic solids. This is shown in Figure \ref{fig:tiles} (see also \citep{so_3_voronoi}). In this figure the fundamental domains $F_{H \backslash \textnormal{SO}(3)}$ are depicted in exponential coordinates in $\mathfrak{so}(3)$ (identified with a ball of radius $\pi$, as explained above). Note that this is a conceptual plot, since actually the edges and faces of these Voronoi cells are slightly bent. The number $|H|$ of elements in $H$ is 12 for the group of tetrahedral, $24$ for the group of octahedral, and $60$ for the group of icosahedral rotational symmetries. By the left (or right) action of $H$ on the respective fundamental domain, it is possible to (almost completely) cover $\textnormal{SO}(3)$, cf.\ \eqref{eq:right_coset_fundamental_decomposition}.

\begin{figure}[t]
\centering
\includegraphics[width=0.32\textwidth]{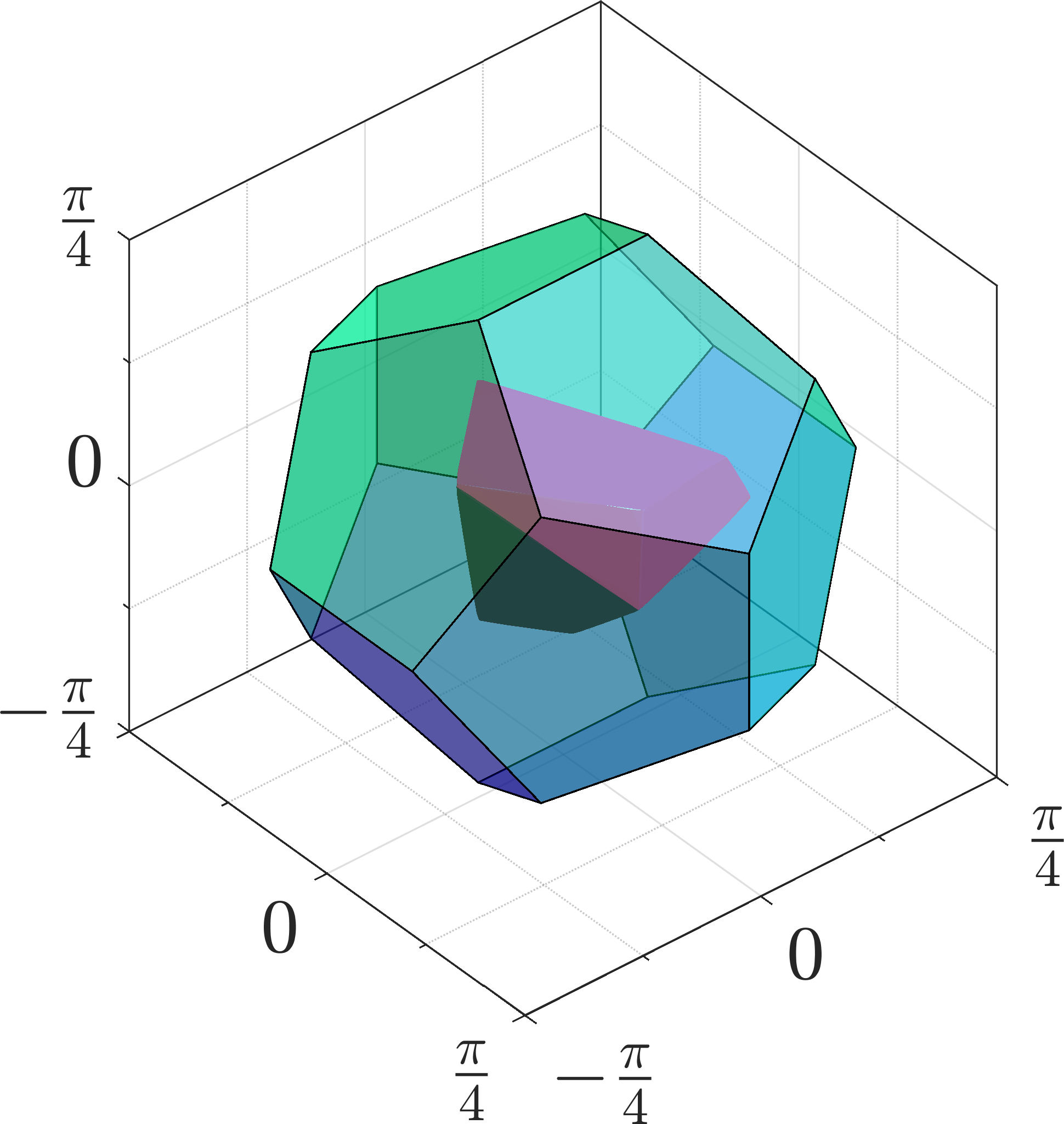}\hfill
\includegraphics[width=0.32\textwidth]{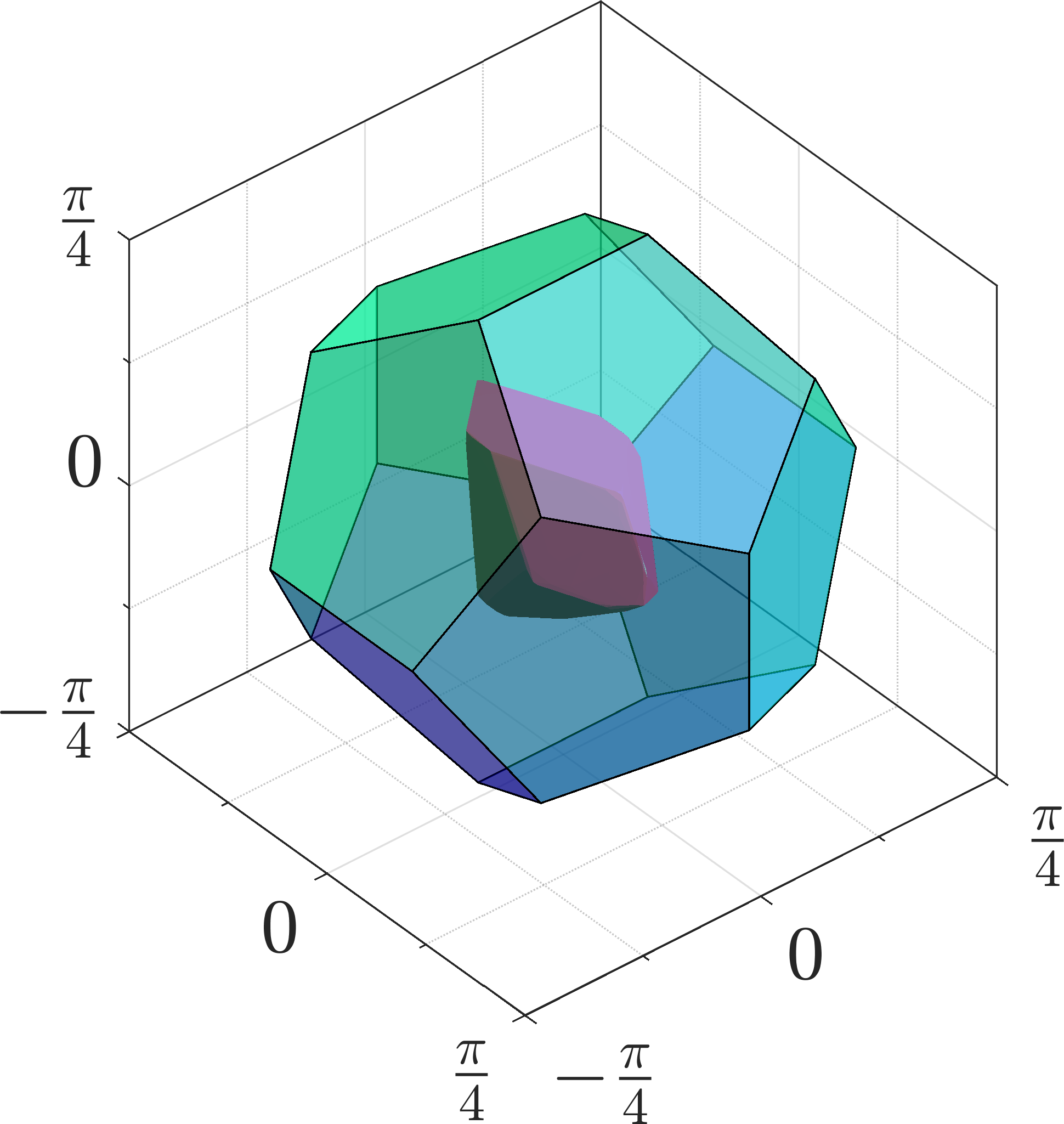}\hfill
\includegraphics[width=0.32\textwidth]{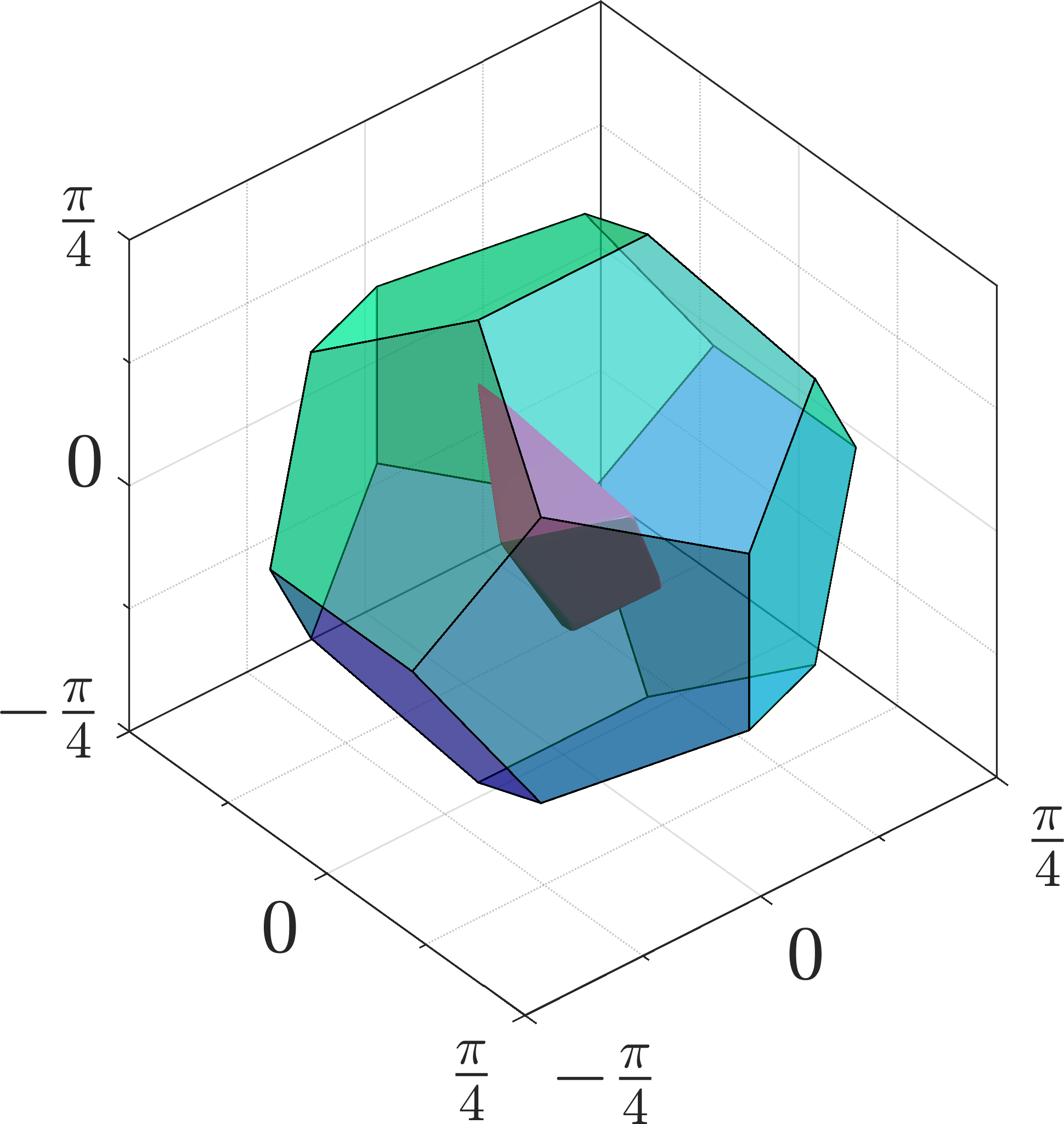}
\caption{Center Voronoi cell $F_{H \backslash \textnormal{SO}(3)}$ in coset space (emerald region) with $H$ as the group of icosahedral symmetries, and center Voronoi cell $F_{H \backslash \textnormal{SO}(3) / K}$ in double-coset space (ruby region) with $H$ as the group of icosahedral symmetries for all cases and $K$ as a conjugated group of (from left to right) tetrahedral, octahedral, and icosahedral symmetries, respectively.}
\label{fig:shards}
\end{figure}

\begin{figure}[t]
\centering
\includegraphics[width=0.4\textwidth]{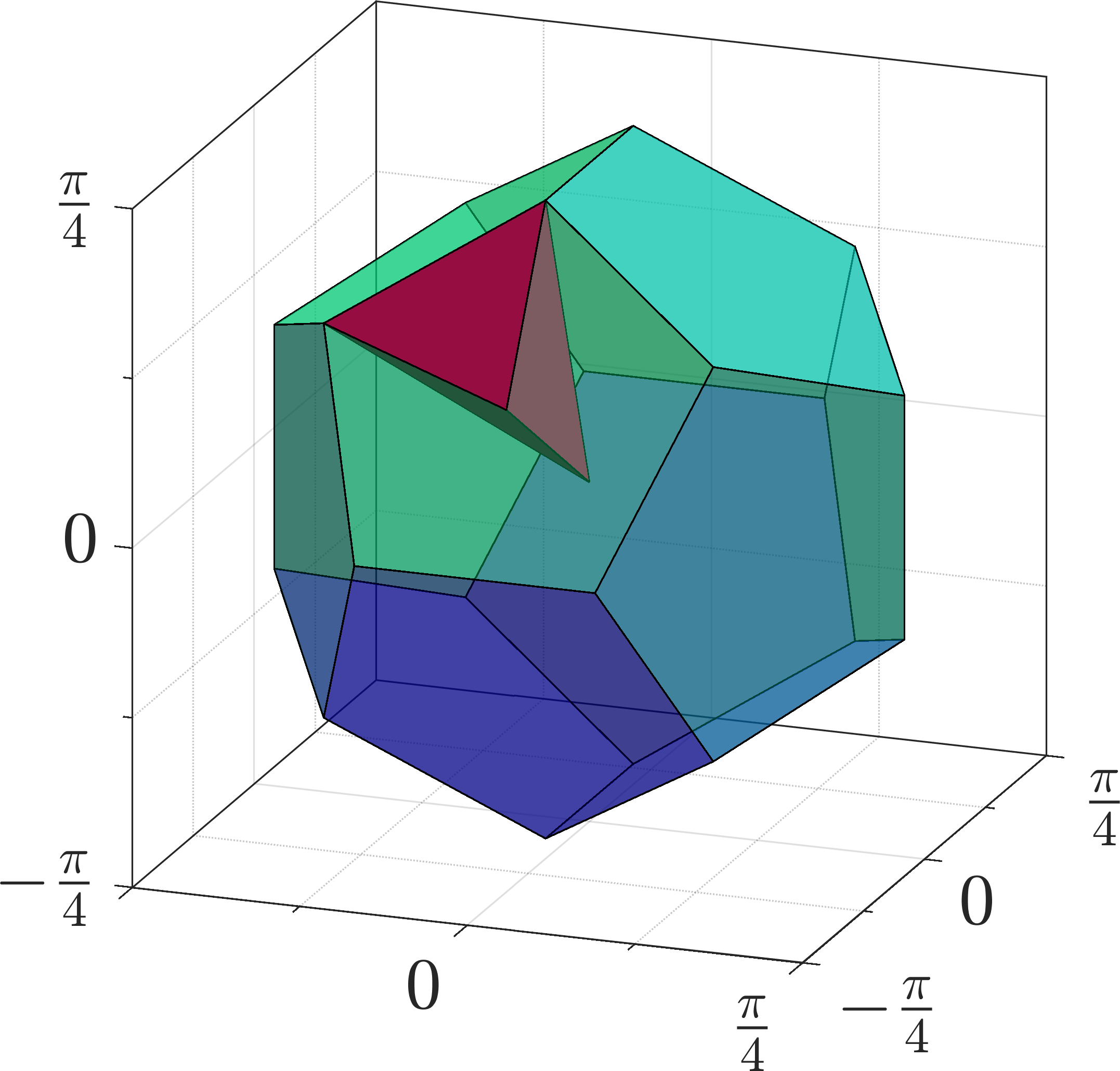}
\caption{Dodecahedral cell $F_{H \backslash \textnormal{SO}(3)}$ (emerald region) and tetrahedral wedge $F_{H \backslash \textnormal{SO}(3) / H}$ (ruby region), with $H$ as the group of icosahedral symmetry. The dodecahedral cell can be decomposed into 60 identical tetrahedral wedges, with five packed on each pentagonal face.}
\label{fig:wedge}
\end{figure}

If $H$ is the group of rotational symmetry operations of the icosahedron, then $|H| = 60$ and $F_{H \backslash \textnormal{SO}(3)}$ can be viewed as a dodecahedral cell centered at the origin of the Lie algebra $\mathfrak{so}(3)$ (see Fig.\,\ref{fig:tiles}, right). If we choose the second subgroup $K$ to be a conjugated group of the tetrahedral, octahedral, or icosahedral symmetries (\textit{i.\,e.}, $K \coloneqq gPg^\textnormal{T}$, where $P$ is the group of the rotational symmetries of the respective Platonic solid and $g \in \textnormal{SO}(3)$ is chosen such that $H \cap K = \{\mathbb{I}\}$), then the Voronoi cell $F_{H \backslash \textnormal{SO}(3) / K}$ takes a shape as exemplarily shown in Figure \ref{fig:shards}. On the other hand, if we choose $K \coloneqq H$, then $F_{H \backslash \textnormal{SO}(3) / H}$ cannot be constructed as a Voronoi cell, but it can be chosen as an irregular tetrahedron (the ruby region in Fig.\,\ref{fig:wedge}), yielding a 
subdivision of the dodecahedral cell $F_{H \backslash \textnormal{SO}(3)}$ by conjugation of the tetrahedron with the elements in $H$. Note that such conjugation has the effect of rotation in $\mathfrak{so}(3)$ since $\log^\vee(g R g^\textnormal{T}) = g \log^\vee\! R$. Similarly, if $K < H$, then a $|K|$-fold division of $F_{H \backslash \textnormal{SO}(3)}$ can be computed to represent $F_{H \backslash \textnormal{SO}(3) / K}$, and $F_{H \backslash \textnormal{SO}(3)}$ can be reconstructed from these pieces, again by conjugation.

When choosing $H$ to be icosahedral and $K = H$ or $K = gHg^\textnormal{T}$ this means that we can divide $\textnormal{SO}(3)$ into 3600 pieces of equal size. 
The 3600 respective barycentric or Voronoi centers of these cells can be taken as a discretization of $\textnormal{SO}(3)$, and any of these centers can be written in a unique way as $R_{ij} = h_i k_j$ where $(h_i,k_j) \in H \times K$ with $i,j \in \{1,2,...,60\}$. This means that any one of the 3600 points $R_{ij}$ corresponds to a two-letter word $(h_i,k_j)$.

A natural question to ask is then, for $R \in \textnormal{SO}(3)$, how do we find the closest word $(h_i,k_j)$ to approximate it. This decoding, or ``signals to symbols'' problem is addressed in Section \ref{subsec:rotation_decoding}.

\section{Rigid-body motions as a group used for framing robots}
\label{sec:rigid_body_motions}
Given any rigid object or multi-rigid-body actor (such as a human, robot, self-driving car, or smart house), the behavior and use of this item involves the time evolution of its conformation and its overall position and orientation, or `pose'. A common descriptive framework for both the internal (conformational) degrees of freedom and their relative pose (configuration) is to attach reference frames on each 
rigid component.

Let $g = (R,\bm{t})$ denote a rigid-body motion relative to a reference frame fixed in space, where $R \in \textnormal{SO}(3)$ is a rotation matrix and $\bm{t} \in \mathbb{R}^3$ is a translation vector. The set of all such motions forms a six-dimensional Lie group, the \emph{special Euclidean group} $\textnormal{SE}(3)$.
This is a group because the composition operation
\begin{equation*}
g_1 g_2 \,=\, (R_1,\bm{t}_1) (R_2,\bm{t}_2) \,\coloneqq\, (R_1 R_2, R_1 \bm{t}_2 + \bm{t}_1)
\end{equation*}
satisfies the properties of closure and associativity, the identity exists and is simply $e = (\mathbb{I},\bm{0})$ (with zero translation), and the inverse of $g$ is $g^{-1} = (R^\textnormal{T}, -R^\textnormal{T} \bm{t})$. Note that $\textnormal{SE}(3)$ is not commutative (Abelian).
The group operation is the same as 
multiplying homogeneous transformation matrices, \textit{i.\,e.},
\begin{equation*}
H(g_1 g_2) \,=\, H(g_1) H(g_2), \quad \textnormal{where} \quad H(g) \,\coloneqq\, 
\begin{bmatrix}
\begin{array}{lc}
R & \bm{t} \\
\bm{0}^\textnormal{T} & 1 
\end{array}
\end{bmatrix}\!.
\end{equation*}

In the case of \emph{planar} motions, we deal with the special Euclidean group $\textnormal{SE}(2)$, where $R \in \textnormal{SO}(2)$ is parameterized by a single angle $\theta$, and $\bm{t} \in \mathbb{R}^2$ has components $(x,y)$, totalling three degrees of freedom. $\textnormal{SO}(2)$ is the group of rotations in $\mathbb{R}^2$, defined analogously to \eqref{eq:SO(3)}.
In general, $\textnormal{SE}(d)$ is an example of an \emph{outer} (or \emph{external}) \emph{semi-direct product} which combines the two groups $(\mathbb{R}^d,+)$ and $\textnormal{SO}(d)$ into the new group
\begin{equation*}
\textnormal{SE}(d) \,\coloneqq\, \textnormal{SO}(d) \ltimes \mathbb{R}^d.
\end{equation*}
The underlying set of this group is the Cartesian product $\textnormal{SO}(d) \times \mathbb{R}^d$, but the symbol $\ltimes$ reflects the fact that the group operation is not simply $(R_1, \bm{t}_1)(R_2, \bm{t}_2) \,=\, (R_1 R_2, \bm{t}_1 + \bm{t}_2)$, which is also a group (called the \emph{direct product}), but does not reflect the way that rigid-body motions work.


The next section reviews the 
handedness-preserving (Sohncke) crystallographic space groups, which are discrete subgroups of $\textnormal{SE}(3)$ and which form an important component of the motion alphabets developed in this paper.

\section{Crystallographic groups}
\label{sec:crystallographic_groups}
A \emph{crystallographic group} is a discrete (and also co-compact) subgroup of the \emph{Euclidean group} $\textnormal{E}(d) \coloneqq \textnormal{O}(d) \ltimes \mathbb{R}^d$, 
where $\textnormal{O}(d)$ is the \emph{orthogonal group} 
consisting of all orthogonal real-valued $d \times d$ matrices (defined as in \eqref{eq:SO(3)} for $d = 3$, but also allowing $\det R = -1$ there). In addition to rotations, the group $\textnormal{O}(d)$ also contains \emph{reflections} and \emph{roto-reflections} (\emph{improper rotations}). 
If $d = 3$ a crystallographic group is commonly called a \emph{space group}, for $d = 2$ it is referred to as a \emph{wallpaper group}. The literature on mathematical crystallography is immense, and spans many decades. Some notable introductions include \citep{janssen,Hahn,aroyo1,VolA1}. The relationship between torsion-free crystallographic groups (\textit{i.\,e.}, Bieberbach groups, see below) and flat manifolds has been studied extensively \citep{Hantzsche,charlap,montesinos,wolf,sepanski}.

Elements of a crystallographic group $\Gamma$ can be expressed as pairs
\begin{equation}\label{eq:crystallographic_element}
\gamma \,=\, (R_\gamma, \bm{t}_\gamma + \bm{v}(R_\gamma)),
\end{equation}
where $R_\gamma \in \mathbb{P}$ (a discrete \emph{point group}, \textit{i.\,e.}, a subgroup of $\textnormal{O}(d)$), $\bm{t}_\gamma \in \mathbb{L}$ (a lattice in $\mathbb{R}^d$), and $\bm{v} : \mathbb{P} \to \mathbb{R}^d$. In particular, $\bm{v}$ is the translational part of a glide-reflection or screw-displacement lattice motion. In general $\bm{v}$ will satisfy the \emph{co-cycle identities}
\begin{align*}
\bm{v}(\mathbb{I}) \,&=\, \bm{0},\\
\bm{v}(R_{\gamma_1} R_{\gamma_2}) \,&=\, (R_{\gamma_1} \bm{v}(R_{\gamma_2}) + \bm{v}(R_{\gamma_1})) \, \textnormal{mod} \, T,
\end{align*}
where $T \coloneqq \{\mathbb{I}\} \ltimes \mathbb{L}$ is the subgroup of pure (or \emph{primitive}) translations in $\Gamma$, which is always normal ($T \triangleleft \Gamma$). The ``$\textnormal{mod} \, T$'' removes components in the sum that are in $T$, in analogy with for example $(1 + 5) \,\textnormal{mod}\, 4 = 2$ in modulo-$4$ arithmetic.

If an element $\gamma \in \Gamma \setminus \{e\}$ is of \emph{finite order} (\textit{i.\,e.}, if there exists an $n \in \mathbb{N}$ such that $\gamma^n = e$), it 
is called a \emph{torsion element}. The group $\Gamma$ is called \emph{torsion-free} (or a \emph{Bieberbach group}) if it is free of torsion elements. 
This is equivalent to the property that no element $\gamma \in \Gamma$ other than the identity $e$ has a fixed point (\textit{i.\,e.}, a point $\bm{p} \in \mathbb{R}^d$ with $\gamma \bm{p} = \bm{p}$).
If 
$\bm{v} \equiv \bm{0}$ 
in \eqref{eq:crystallographic_element}, then $\Gamma$ can be written 
as the 
semi-direct product $\Gamma = \mathbb{P} \ltimes \mathbb{L}$ and the group is called \emph{symmorphic}. 
Of the 230 possible types of space groups, 73 can be decomposed in this way. These are the \emph{symmorphic space groups}. 
Bieberbach groups are not symmorphic. 

In the theory of crystallographic groups, a well-known isomorphism is due to the above-mentioned fact that the translation subgroup $T$ of $\Gamma$ is normal:
\begin{equation*}
\frac{\Gamma}{T} \,\cong\, \mathbb{P}.
\end{equation*}
Moreover, it can be shown that $\mathbb{P}$ is not only discrete, but must be \emph{finite}. In the case $d = 3$ the point group $\mathbb{P}$ will belong to one of 32 discrete \emph{crystallographic point groups} that constitute the so-called \emph{crystal classes}. 

In the context of space groups, we can distinguish between Bieberbach (\textit{i.\,e.}, torsion-free) groups as one extreme, 
and groups which contain only the identity and torsion elements (rotations, reflections, and improper rotations) as the other extreme, all other space groups lying somewhat ``in between.'' 
For a symmorphic space group $\Gamma$, a Bieberbach subgroup $\Gamma_\textnormal{B} < \Gamma$ with minimum index in $\Gamma$ is the subgroup $T$ of primitive translations; 
for many non-symmorphic space groups, on the other hand, there is a Bieberbach subgroup $\Gamma_\textnormal{B}$ with index $[\Gamma : \Gamma_\textnormal{B}] < [\Gamma : T]$ allowing for a decomposition of $\Gamma$ as 
a group 
product \citep[p.\,719]{ratnay}
\begin{equation}\label{eq:semi1}
\Gamma \,=\, \Gamma_\textnormal{B} S \,\coloneqq\, \{\gamma_\textnormal{B} s \,:\, \gamma_\textnormal{B} \in \Gamma_\textnormal{B}, \, s \in S\}, 
\end{equation}
where $S < \Gamma$ is a proper subgroup of $\mathbb{P} \ltimes \{\bm{0}\} < \textnormal{E}(3)$, and thus $\Gamma_\textnormal{B} \cap S = \{e\}$. If $\Gamma_\textnormal{B} \triangleleft \Gamma$ then \eqref{eq:semi1} is an \emph{inner} (or \emph{internal}) semi-direct product ($\Gamma = \Gamma_\textnormal{B} \rtimes S$). In this case, the decomposition $\gamma = \gamma_\textnormal{B} s$ is unique for each $\gamma \in \Gamma$.
Another useful property is that each of the 65 \emph{Sohncke space groups} (\textit{i.\,e.}, handedness-preserving space groups $\Gamma < 
\textnormal{SE}(3)$) can be written as 
a 
product
\begin{equation}\label{eq:semi2}
\Gamma = \Gamma_\textnormal{B} 
\Gamma_\textnormal{S} \,\coloneqq\, \{\gamma_\textnormal{B} \gamma_\textnormal{S} \,:\, \gamma_\textnormal{B} \in \Gamma_\textnormal{B}, \, \gamma_\textnormal{S} \in \Gamma_\textnormal{S}\}, 
\end{equation}
where $\Gamma_\textnormal{B}$ and $\Gamma_\textnormal{S}$ are respectively Bieberbach and symmorphic subgroups 
(see \citep[Thm.\,3]{ratnay}). If $\Gamma_\textnormal{B} \cap \Gamma_\textnormal{S} = \{e\}$ and $\Gamma_\textnormal{B}$ or $\Gamma_\textnormal{S}$ is normal, then 
\eqref{eq:semi2} is
an inner semi-direct product, and for every $\gamma \in \Gamma$ there exist unique $\gamma_\textnormal{B} 
\in \Gamma_\textnormal{B}$ and $\gamma_\textnormal{S} 
\in \Gamma_\textnormal{S}$ such that $\gamma = \gamma_\textnormal{B} \gamma_\textnormal{S}$. 
If both $\Gamma_\textnormal{B}$ and $\Gamma_\textnormal{S}$ are normal, then $\Gamma$ is decomposed into the inner direct product $\Gamma = \Gamma_\textnormal{B} \times \Gamma_\textnormal{S}$. 

\section{A motion alphabet based on a fine double-coset decomposition}
\label{sec:motion_alphabet}
The essence of language is communicating information about the continuous world using a finite number of symbols, characters, or patterns. In the same spirit, in classical AI, a fundamental goal is to convert perceptual information from the continuous world to strings of symbols drawn from a finite alphabet, so that the machinery of finite-state automata reasoning can be applied. Every action of a robot in the physical world can be described as a sequence of paths 
in $\textnormal{SE}(3)$. 
Such paths are continuous mathematical trajectories which can be viewed as `signals'.

A fundamental problem in AI and sensory perception is the so-called ``signals to symbols'' problem \citep{Jackendoff,Winograd,Russell} in which observations in the continuous world are converted to coarsified representations (\textit{i.\,e.}, symbols) on which AI systems can execute logical reasoning algorithms. A natural way to discretize the space of paths is to first discretize time, thereby reducing the infinite dimensions of the path space to a finite number of dimensions, and then to replace each sampled pose on the path with a rounded-off version in a discrete set (the alphabet of maneuvers). Because $\textnormal{SE}(3)$ does not have dense subgroups, a clever discretization needs to be constructed.

In the three-dimensional case, one of the finest space groups is $\Gamma = \textnormal{P432}$ which has a total of 24 rotational elements corresponding to the rotational symmetry operations of a cube (cf.\ \citep[p.\,634f]{Hahn}). Therefore, if we want to quantize rigid-body motions with a resolution that is reasonable for real-world tasks, something finer than rounding off to the nearest element of $\Gamma$ is necessary.

Building on ideas introduced in the context of protein-packing models in X-ray crystallography \citep{chirik-acta1,yy2013,gc-xstal1,ratnay,acta4}, we can augment $\Gamma$ by an auxiliary discrete rotation group. For example, let $\Delta$ denote the group of rotational symmetries of the icosahedron (as a discrete subgroup of $\textnormal{SE}(3)$). $\Delta$ has 60 elements, and there is no lattice of discrete translations that corresponds to it. In particular, $\Gamma, \Delta < \textnormal{SE}(3)$ and $\Gamma \cap \Delta = \{e\}$, the identity element. This means that the double-coset space $\Gamma \backslash \textnormal{SE}(3) / \Delta$ is a compact Riemannian manifold. It is possible to define a compact fundamental domain $F_{\Gamma \backslash \textnormal{SE}(3) / \Delta} \subset \textnormal{SE}(3)$ using \eqref{eq:pre_Voronoi_double_coset}. Restricting the quotient map from $\textnormal{SE}(3)$ to $\Gamma \backslash \textnormal{SE}(3) / \Delta$
%
%
to the fundamental domain $F_{\Gamma \backslash \textnormal{SE}(3) / \Delta}$ then gives a bijective (\textit{i.\,e.}, one-to-one) mapping from $F_{\Gamma \backslash \textnormal{SE}(3) / \Delta}$ to $\Gamma \backslash \textnormal{SE}(3) / \Delta$. 
Moreover, the action of $\Gamma$ (from the left) and $\Delta$ (from the right) gives a way to tile $\textnormal{SE}(3)$ with disjoint 
shifts of 
$F_{\Gamma \backslash \textnormal{SE}(3) / \Delta}$, because (cf.\ \eqref{eq:double_coset_fundamental_decomposition})
\begin{equation}\label{eq:tilingeq}
\textnormal{SE}(3) \,=\, \bigcup_{\gamma \in \Gamma} \bigcup_{\delta \in \Delta} \gamma F_{\Gamma \backslash \textnormal{SE}(3) / \Delta} \delta.
\end{equation}
Intensive study of the fundamental domains $F_{\Gamma \backslash \textnormal{SE}(3)}$ and $F_{\Gamma \backslash \textnormal{SE}(3) / \Delta}$ has been conducted (\textit{ibid.}).

\begin{figure}[t]
\centering
\includegraphics[width=\textwidth]{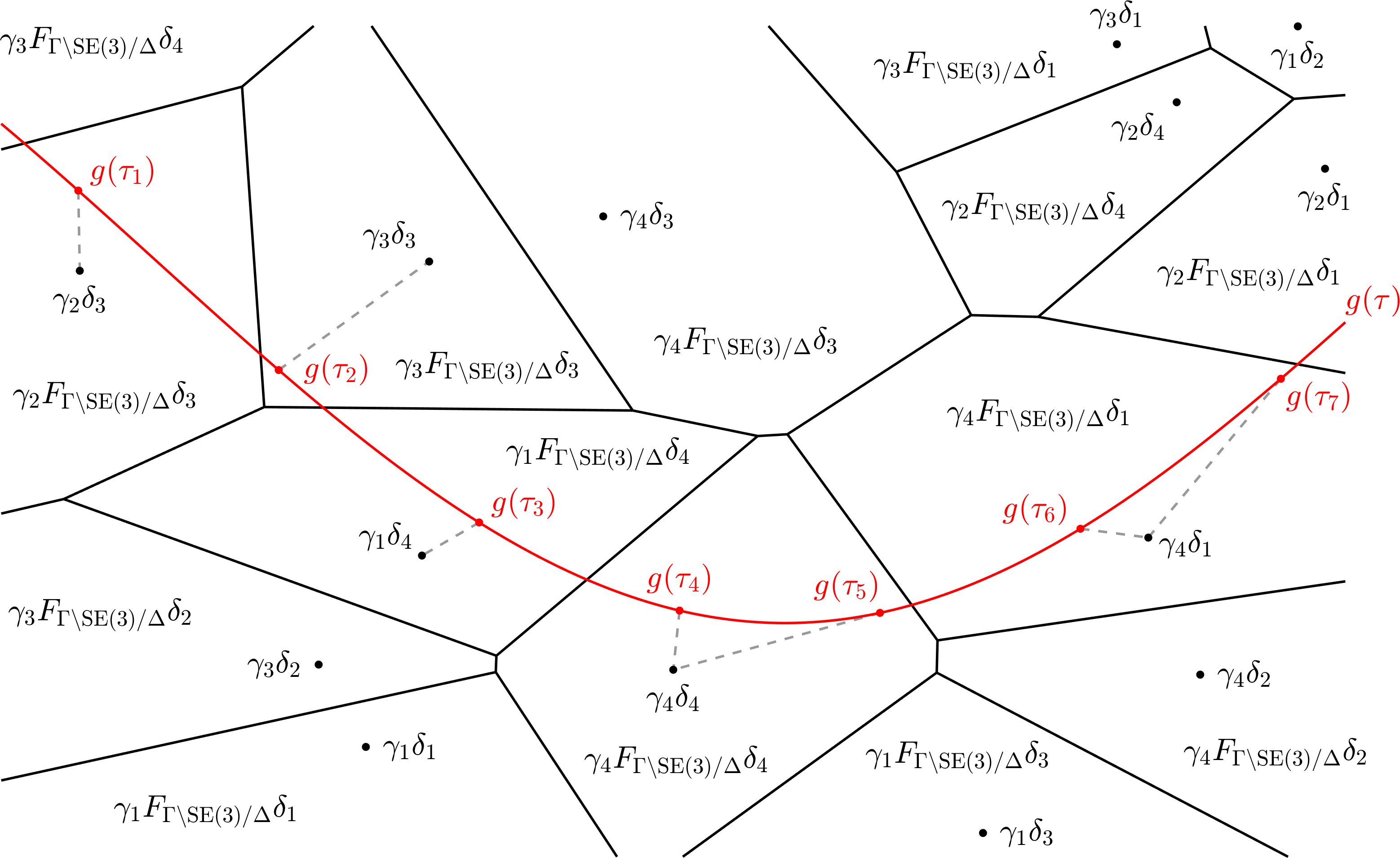}
\caption{Discretizing a continuous motion trajectory $g$ at times $\tau_1,\dots,\tau_7$ using the alphabet $\Gamma \times \Delta$ (conceptual plot). After discretization the continuous motion can be expressed as the sentence $(\gamma_2,\delta_3)$, $(\gamma_3,\delta_3)$, $(\gamma_1,\delta_4)$, $(\gamma_4,\delta_4)$, $(\gamma_4,\delta_4)$, $(\gamma_4,\delta_1)$, $(\gamma_4,\delta_1)$.}
\label{fig:conceptual_plot}
\end{figure}

When the fundamental domain $F_{\Gamma\backslash \textnormal{SE}(3)/\Delta}$ is constructed using \eqref{eq:pre_Voronoi_double_coset}, it has the identity element $e$ at its center, and so the tiling in \eqref{eq:tilingeq} has the effect of sampling each center point by moving from $e$ to $\gamma \delta$ where $\gamma \in \Gamma$ and $\delta \in \Delta$. In other words, the product $\Gamma \times \Delta < \textnormal{SE}(3)^2$ can be used as a quantized version of $\textnormal{SE}(3)$. The number of rotational elements will be $24 \times 60 = 1440$ which is sufficiently fine to capture the essence of any frame along a trajectory during a robot task.
The alphabet defined by $\Gamma \times \Delta$ is infinite, but by limiting the extent of translations to be contained in a bounded region, it becomes finite. This means that continuous trajectories can be translated into a finite string of alphabet characters (Fig.\,\ref{fig:conceptual_plot}). This opens up the possibility to map these quantized trajectories into words expressed in a natural language.

An important advantage of the quantization scheme \eqref{eq:tilingeq} is that the shifted fundamental domains $\gamma F_{\Gamma \backslash \textnormal{SE}(3) / \Delta} \delta$ will all have the same volume, \textit{i.\,e.},
\begin{equation}\label{eq:same_volume}
\mu(\gamma F_{\Gamma \backslash \textnormal{SE}(3) / \Delta} \delta) \,=\, \mu(F_{\Gamma \backslash \textnormal{SE}(3) / \Delta}) \quad\textnormal{for all}\quad (\gamma,\delta) \in \Gamma \times \Delta,
\end{equation}
where $\mu$ is the (left- and right-invariant) Haar measure on the (unimodular) Lie group $\textnormal{SE}(3)$. This means that the centers $\gamma \delta$ of these shifted fundamental domains used for quantization at the same time also allow for a very uniform sampling of the group $\textnormal{SE}(3)$.

\section{More alphabets and coarse-to-fine decoding algorithms}
\label{sec:decoding_algorithms}
As explained above, after wisely designing a pair of discrete subgroups $\Gamma, \Delta < \textnormal{SE}(3)$ with $\Gamma \cap \Delta = \{e\}$ we can construct a fundamental domain as in \eqref{eq:pre_Voronoi_double_coset}, and decompose $\Gamma$ as in \eqref{eq:tilingeq}.
There are many other ways to choose $F_{\Gamma \backslash \textnormal{SE}(3)}$ and $F_{\Gamma \backslash \textnormal{SE}(3) / \Delta}$, as explained in \citep{acta4}. A particularly simple choice is the Cartesian product
\begin{equation}\label{bigchoice}
F_{\Gamma \backslash \textnormal{SE}(3) / \Delta} \,\coloneqq\, F_{\mathbb{P} \backslash \textnormal{SO}(3) / \Delta} \,\times\, F_{\mathbb{L} \backslash \mathbb{R}^3},
\end{equation}
where $\mathbb{P} \cong \frac{\Gamma}{T}$ is the point group of $\Gamma$ and $\mathbb{L}$ the lattice of primitive translations. 
Alternatively, given a decomposition \eqref{eq:semi1}, another natural choice is
\begin{equation*}
F_{\Gamma \backslash \textnormal{SE}(3) / \Delta} \,\coloneqq\, F_{S \backslash \textnormal{SO}(3) / \Delta} \,\times\, F_{\Gamma_B \backslash \mathbb{R}^3}.
\end{equation*}
Here $S \backslash \textnormal{SO}(3) / \Delta$ and $\Gamma_B \backslash \mathbb{R}^3$ are not (double-)coset spaces\,--\,because $S \nless \textnormal{SO}(3)$ and $\Gamma_\textnormal{B} \nless \mathbb{R}^3$\,--\,but are \emph{orbit spaces} consisting respectively of \emph{orbits} $S R \Delta \coloneqq \{s R \delta : s \in S, \, \delta \in \Delta\}$ and $\Gamma_B \bm{x} \coloneqq \{\gamma_\textnormal{B} \bm{x} : \gamma_\textnormal{B} \in \Gamma_\textnormal{B}\}$ ($R \in \textnormal{SO}(3)$, $\bm{x} \in \mathbb{R}^3$). The fundamental domains $F_{S \backslash \textnormal{SO}(3) / \Delta}$ and $F_{\Gamma_B \backslash \mathbb{R}^3}$ above can be constructed by choosing exactly one point per orbit, which is consistent with the definitions in Section \ref{sec:group_theory}. Different fundamental domains $F_{\Gamma \backslash \textnormal{SE}(3) / \Delta}$ such as those above can be used to express different quantizations via \eqref{eq:tilingeq}.

\subsection{The purely rotational case}\label{subsec:rotation_decoding}

The choice for $F_{\Gamma \backslash \textnormal{SE}(3) / \Delta}$ in \eqref{bigchoice} allows us to bootstrap off of the fundamental domains for double-coset spaces for $\textnormal{SO}(3)$ discussed earlier. In fact, we can go even further and describe an $\textnormal{SE}(3)$ motion trajectory $g(\tau) = (R(\tau),\bm{t}(\tau))$ in $\textnormal{SO}(3) \times \mathbb{R}^3$ (as a direct product rather than a semi-direct product). This is not merely to make things easier\,--\,viewing pose change trajectories in this way has some advantages, as described in \citep{posechangepaper}, where the direct product $\textnormal{SO}(3) \times \mathbb{R}^3$ is called the \emph{pose change group}, and is denoted as $\textnormal{PCG}(3)$. Therefore, below we describe in some detail how the ``signals to symbols'' problem can be solved efficiently in the purely rotational case.

Since \eqref{bigchoice} is a set rather than a group, we can view it as either a subset of $\textnormal{SE}(3)$ or $\textnormal{PCG}(3)$. Either way, the general decoding problem reduces to this: Given $H,K < \textnormal{SO}(3)$ and $R \in \textnormal{SO}(3)$, how can we efficiently find the unique pair $(h_i, k_j) \in H \times K$ such that
\begin{equation}\label{eq:decomp_R}
R \,=\, h_i Q k_j
\end{equation}
with $Q \in F_{H \backslash \textnormal{SO}(3) / K}$.
%
Especially if the Voronoi choice is made for $F_{H \backslash \textnormal{SO}(3) / K}$,
solving \eqref{eq:decomp_R} allows for simply rounding off $R$ to $h_i k_j$, as indicated in Section \ref{subsec:SO(3)}. 

The question then becomes how to do this. With the crystallographic constraint, in $\textnormal{SE}(3)$ it is possible to define $H$ such that $|H| = 24$ (octahedral symmetry) and $|K| = 60$ (icosahedral symmetry), leading to $24 \times 60 = 1440$ combinations. In $\textnormal{PCG}(3)$, on the other hand, subgroups need not be restricted to the crystallographic constraint, and we can have more rotational elements (see below). The question is, is there a better way to test for $h_i$ and $k_j$ than two nested for loops over $i$ and $j$ resulting in a large number of evaluations to find where $\rho(R, h_i k_j)$ is minimized\,--\,which is equivalent to solving \eqref{eq:decomp_R} when the Voronoi choice is made for $F_{H \backslash \textnormal{SO}(3) / K}$.

The answer is positive, and we shall now present a technique to achieve this. Consider the double-coset space $H \backslash \textnormal{SO}(3) / K$, where $H$ is the group of rotational symmetries of the icosahedron, and $K = gHg^\textnormal{T}$ is a conjugated group, $g$ being chosen so that $H \cap K = \{\mathbb{I}\}$. It is thus $|H \backslash \textnormal{SO}(3) / K| = |H| \times |K| = 60^2 = 3600$. We can construct a fundamental domain $F_{H \backslash \textnormal{SO(3)}}$ for the coset space $H \backslash \textnormal{SO(3)}$ as a dodecahedral Voronoi cell (cf.\ \eqref{eq:SO(3)_coset_Voronoi} and Fig.\,\ref{fig:tiles}, right). Due to the Voronoi property, we can find the shifted tile $h_i F_{H \backslash \textnormal{SO(3)}}$ containing the rotation $R$ of interest by computing the distance $\rho(R,h_i)$ of $R$ to the 60 tile centers $h_i \in H$. We then know that $h_i^\textnormal{T} R$ lies in the identity-centered tile $F_{H \backslash \textnormal{SO(3)}}$. We can construct the fundamental domain $F_{H \backslash \textnormal{SO}(3) / K}$ for the double-coset space as a Voronoi cell, too (cf.\ \eqref{eq:SO(3)_double_coset_Voronoi} and Fig.\,\ref{fig:shards}, right). Since the shifts $h_i F_{H \backslash \textnormal{SO}(3) / K} k_j$ 
of this identity-centered fundamental domain will cover $\textnormal{SO}(3)$, they will also cover $F_{H \backslash \textnormal{SO(3)}}$. But the number of required shifts will be much smaller than $3600$. Indeed, we found that approximately $180$ shifted fundamental domains $h_i F_{H \backslash \textnormal{SO}(3) / K} k_j$ are sufficient when the conjugation element $g$ is empirically chosen such as to minimize the extent
\begin{equation*}
\sup_{R \in F_{H \backslash \textnormal{SO}(3) / K}} \rho (R,\mathbb{I})
\end{equation*}
of $F_{H \backslash \textnormal{SO}(3) / K}$. We chose $g$ in this particular way so as to minimize the round-off error in the quantization scheme. We can now quickly find the shifted fundamental domain $h_{i'} F_{H \backslash \textnormal{SO}(3) / K} k_j$ containing the above $h_i^\textnormal{T} R$ by exploiting the Voronoi property of these shifted domains. We then have that $R \in (h_i h_{i'}) F_{H \backslash \textnormal{SO}(3) / K} k_j$, and so the decomposition \eqref{eq:decomp_R} is found. Instead of the 3600 distances on $\textnormal{SO}(3)$, in the above technique, we have to compute only about $60 + 180 = 240$ distances, resulting in a potential speedup of approximately $3600 / 240 = 15$. This was validated in a Matlab test, where we achieved an average speedup of $14.14$ 
using the above strategy with $1000$ random rotations. The test system here was an x86\_64 Unix system with a $3.4$\,GHz Intel Core i7-6700 CPU and with Matlab R2018b. The total runtime was $0.116$\,s using the fast technique as opposed to $1.634$\,s for the Voronoi gold standard (\textit{i.\,e.}, straight-forward minimization of the distance $\rho(R, h_i k_j)$ with respect to $i$ and $j$). In our implementation we used
\begin{equation*}
\log^\vee\! g \,\coloneqq\, [0.435897435897436, -0.076923076923077, -0.128205128205128]^\textnormal{T}
\end{equation*}
and found that the 181 shifted fundamental domains $h_i F_{H \backslash \textnormal{SO}(3) / K} k_j$ with center $h_i k_j$ closest to the identity are sufficient to cover $F_{H \backslash \textnormal{SO}(3)}$. We further exploited the fact that $\rho(R_1,R_2) = \arccos (\frac{1}{2}[\textnormal{tr}(R_1^\textnormal{T} R_2) - 1])$, the trace wherein can be computed efficiently as
\begin{equation*}
\textnormal{tr}(R_1^\textnormal{T} R_2) \,=\, \sum_{i=1}^3 \sum_{j=1}^3 \, [R_1]_{ij} [R_2]_{ij},
\end{equation*}
as well as the fact that $\arccos x < \arccos y$ if and only if $x > y$.

An even more efficient decoding algorithm can be obtained when considering the double-coset space $H \backslash \textnormal{SO}(3) / H$, where $H$ is the group of rotational icosahedral symmetry. Here again we can use the dodecahedral Voronoi fundamental domain $F_{H \backslash \textnormal{SO(3)}}$ for the coset space $H \backslash \textnormal{SO(3)}$, and find the shifted tile $h_i F_{H \backslash \textnormal{SO(3)}}$ containing the rotation $R$ easily as described above. As a fundamental domain $F_{H \backslash \textnormal{SO(3)} / H}$ for the double-coset space, we can choose the tetrahedral wedge shown in Figure \ref{fig:wedge}. We can find the conjugated wedge $h_j F_{H \backslash \textnormal{SO(3)} / H} h_j^\textnormal{T}$ containing the ``pulled-back'' rotation $h_i^\textnormal{T} R$ by using one of the standard query methods. We then know that $R \in (h_i h_j) F_{H \backslash \textnormal{SO(3)} / H} h_j^\textnormal{T}$, and so \eqref{eq:decomp_R} is solved. In a second Matlab test on the same system with $1000$ random rotations, we found that the above technique yields an average speedup of $37.85 \pm 2.41$ as compared with the Voronoi gold standard, the total runtime now being $0.044$\,s in the fast method, with an average of $(0.044 \pm 0.004)$\,ms per rotation.

We note that as is the case in the previous Section \ref{sec:motion_alphabet} (cf.\ \eqref{eq:same_volume}), the shifts of the fundamental domains $F_{H \backslash \textnormal{SO}(3) / K}$ and $F_{H \backslash \textnormal{SO}(3) / H}$ above all have the same volume, which is an important advantage of the double-coset approach presented in this paper.

\subsection{Planar-motion alphabets based on wallpaper groups}

\begin{figure}[t]
\centering
\includegraphics[width=0.32\textwidth]{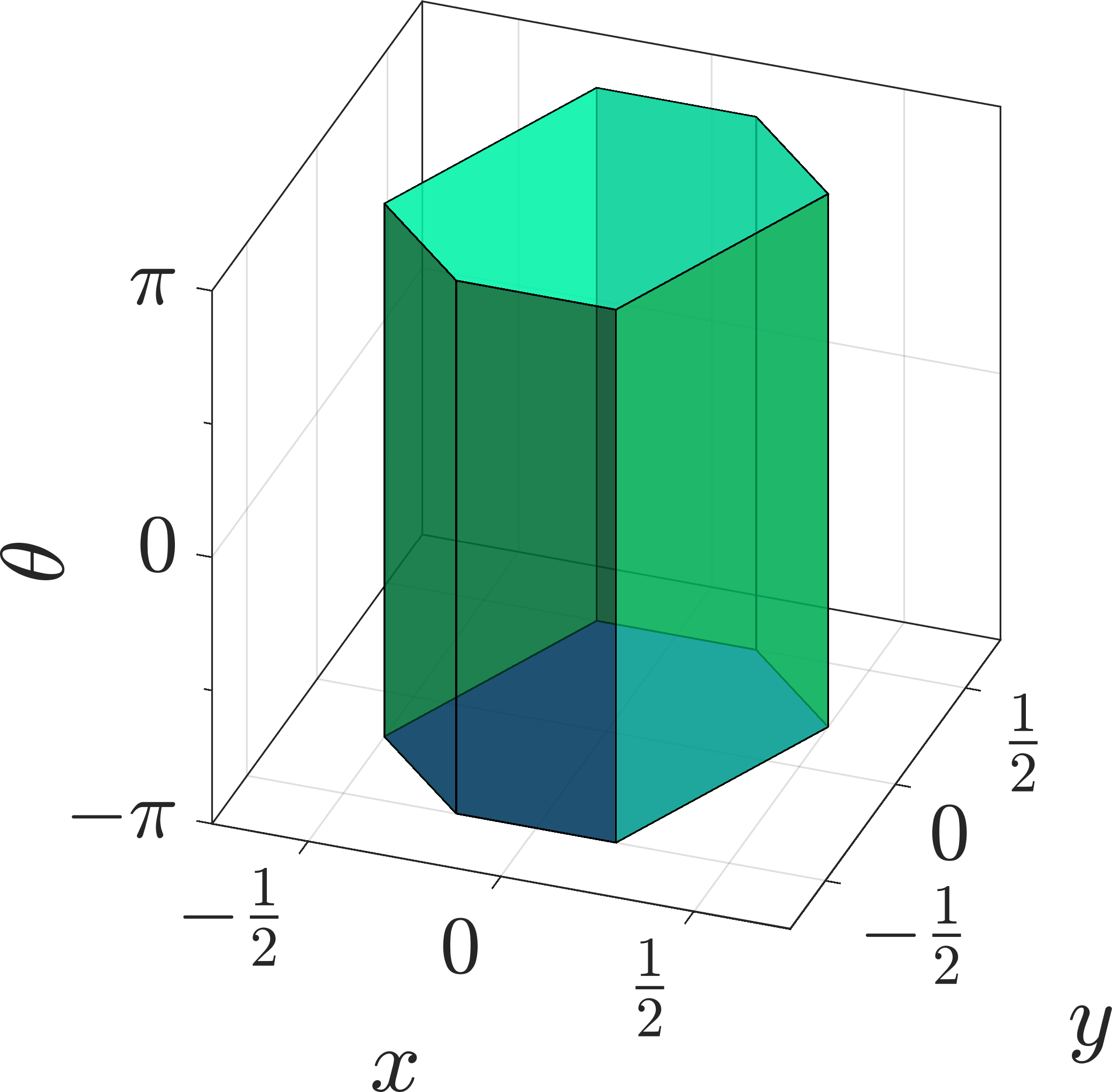}\hfill
\includegraphics[width=0.32\textwidth]{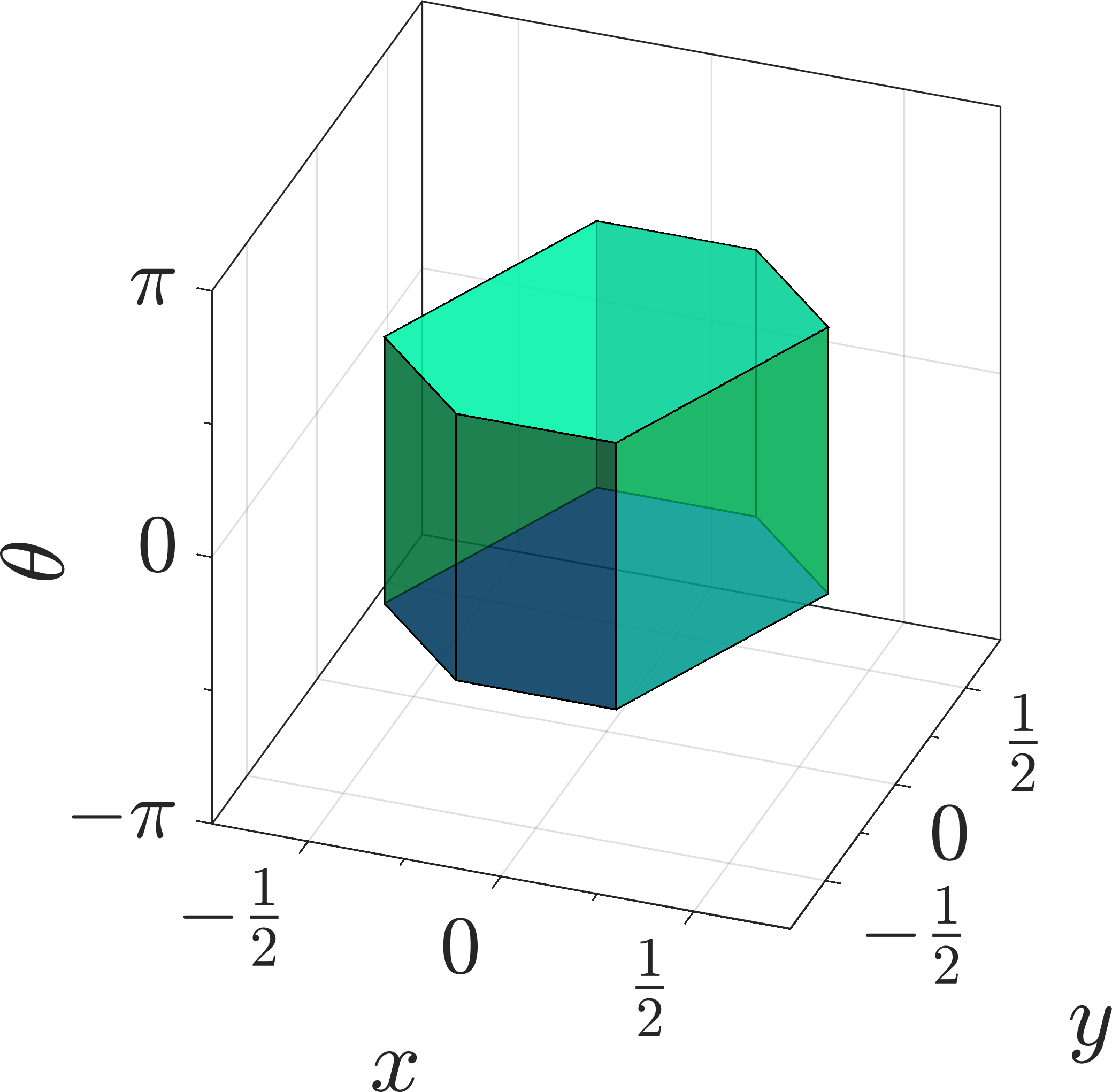}\hfill
\includegraphics[width=0.32\textwidth]{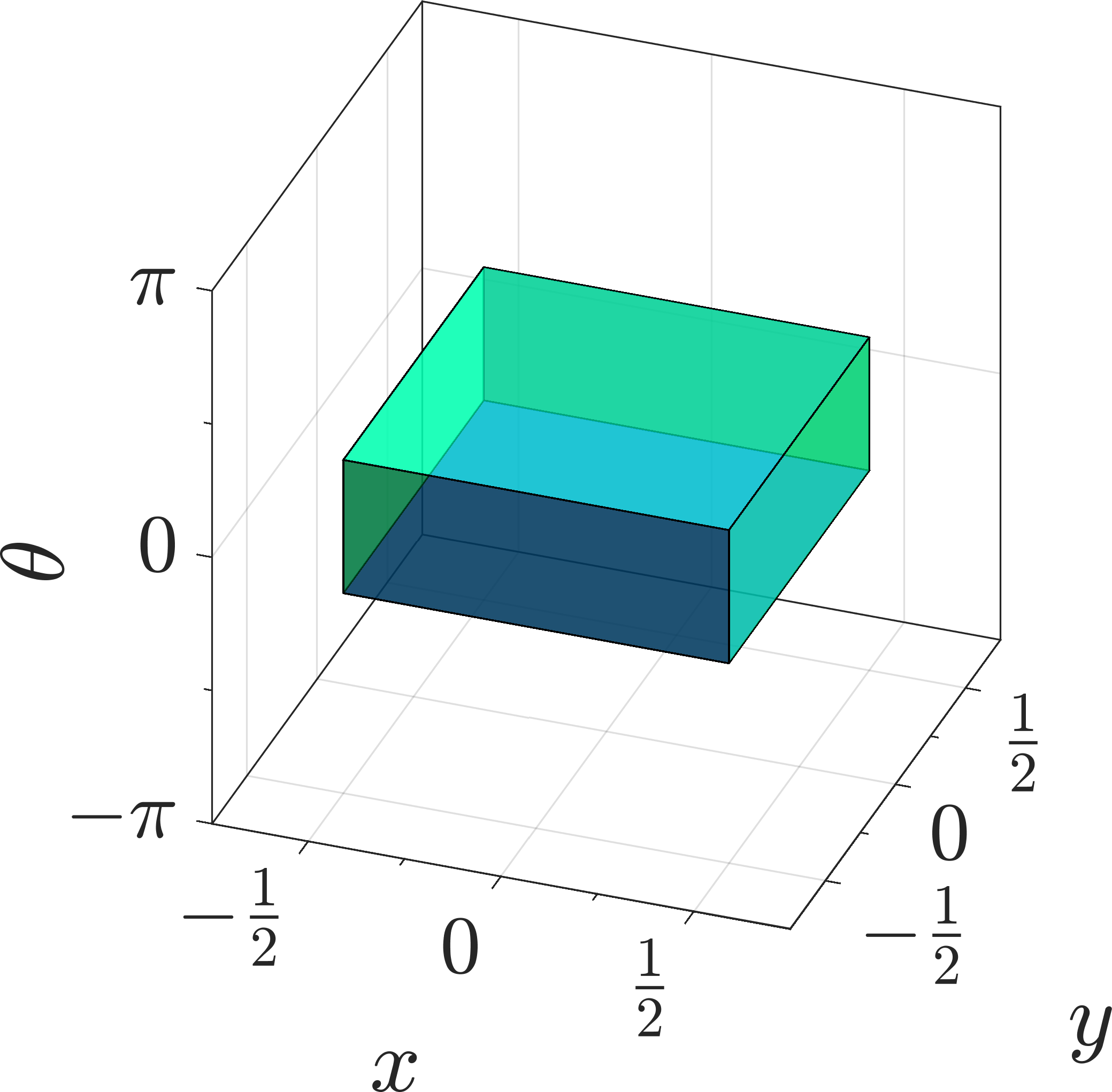}\\\vspace*{11pt}
\includegraphics[width=0.32\textwidth]{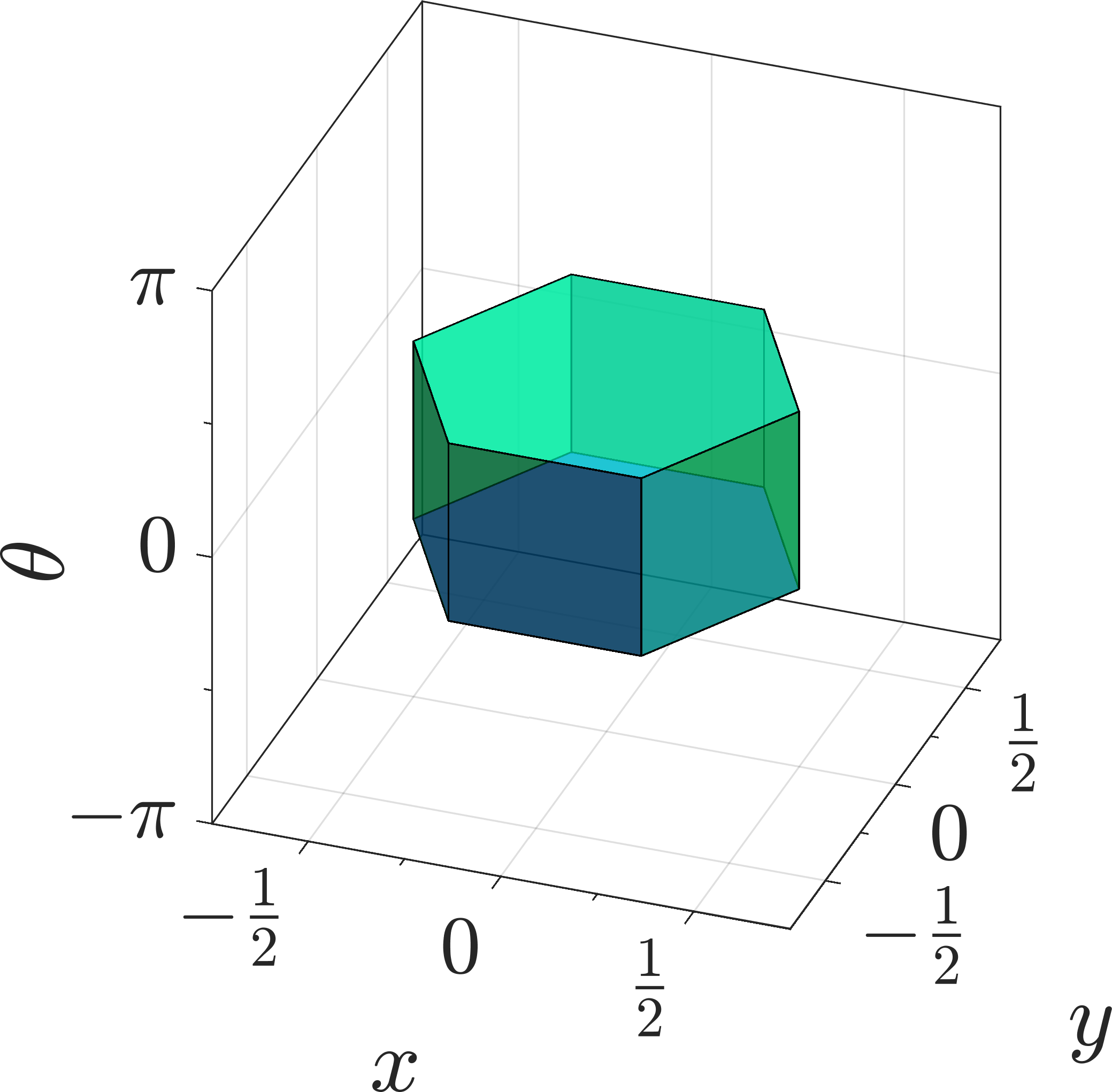}\hspace*{15pt}
\includegraphics[width=0.32\textwidth]{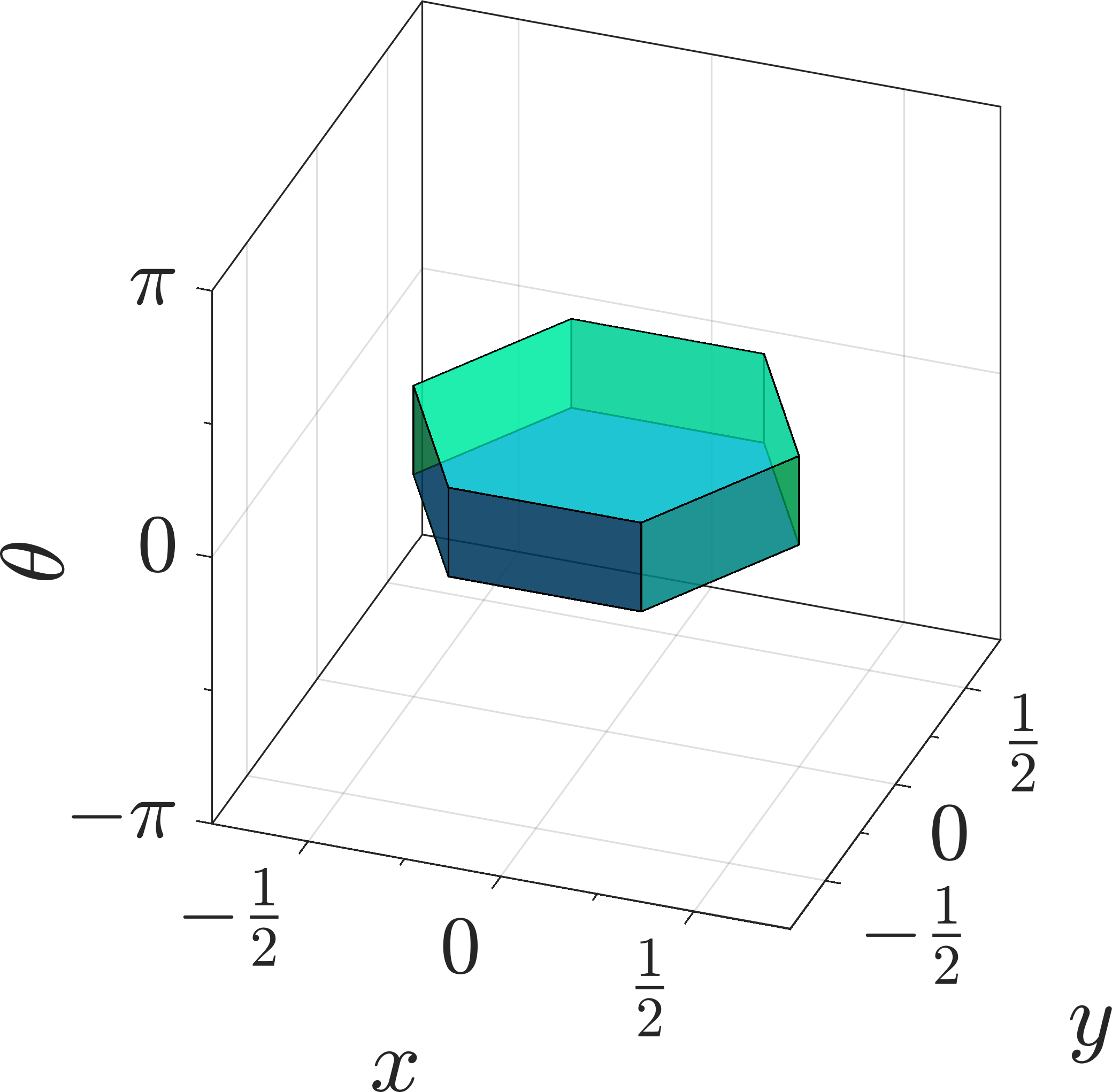}
\caption{Center Voronoi cell $F_{\textnormal{p}_i \backslash \textnormal{SE}(2)}$ for certain instances of the wallpaper groups (left to right, top to bottom) $\textnormal{p}_1$, $\textnormal{p}_2$, $\textnormal{p}_4$, $\textnormal{p}_3$, and $\textnormal{p}_6$.}
\label{fig:AllVoronoiCells}
\end{figure}

As mentioned in Section \ref{sec:crystallographic_groups}, crystallographic groups are called wallpaper groups in the two-dimensional setting. Figure \ref{fig:AllVoronoiCells} shows fundamental domains $F_{\textnormal{p}_i \backslash \textnormal{SE}(2)}$ constructed using \eqref{eq:pre_Voronoi_coset} for instances of the well-known wallpaper groups (cf.\ \citep[Chap.\,6]{Hahn}) $\textnormal{p}_1$, $\textnormal{p}_2$, $\textnormal{p}_4$, $\textnormal{p}_3$, and $\textnormal{p}_6$ (see also \citep{yy2013}), all of them symmorphic. Here $\textnormal{SE}(2)$ is identified with $\mathbb{R}^2 \times (-\pi,\pi) \subset \mathbb{R}^3$, with the $x$ and $y$ axes representing translations in $x$ and $y$ direction and the $z$ axis representing the rotation angle $\theta$. These fundamental domains are generated using the Euclidean metric $\|\cdot - \cdot\|_2$ on $\mathbb{R}^3$, adapted so as to take into account the $2\pi$-periodicity in the rotation angle $\theta$. It is important to note that this metric is left- but not right-invariant (there is no bi-invariant metric on $\textnormal{SE}(2)$). Therefore, the fundamental domains shown in Figure \ref{fig:AllVoronoiCells} are actually Voronoi rather than Voronoi-like cells (as is the case for $\textnormal{SO}(3)$, cf.\ \eqref{eq:SO(3)_coset_Voronoi}).

The group $\textnormal{p}_1$ consists solely of translations, constituting a parallelogrammatic lattice in the translational $x$-$y$ plane. This results in a box with (irregular) hexagonal shape in the $x$-$y$ plane and height $2\pi$ as the fundamental domain $F_{\textnormal{p}_1 \backslash \textnormal{SE}(2)}$. In addition to the translations in $\textnormal{p}_1$, the wallpaper group $\textnormal{p}_2$ also contains a rotation of order two (\textit{i.\,e.}, with angle $\pi$). Therefore, the fundamental domain $F_{\textnormal{p}_2 \backslash \textnormal{SE}(2)}$ also has a hexagonal shape in the translational plane, but the height is only $\pi$ (from $-\pi/2$ to $\pi/2$) instead of $2\pi$. The group $\textnormal{p}_4$ is a group with rotations of order four (\textit{i.\,e.}, with angles $\pi/2$, $\pi$, and $3\pi/2$), as well as translations in a square lattice. Thus the fundamental domain $F_{\textnormal{p}_4 \backslash \textnormal{SE}(2)}$ has the shape of a square in the translational plane, its height being $\pi/2$ (from $-\pi/4$ to $\pi/4$). The groups $\textnormal{p}_3$ and $\textnormal{p}_6$ both have a hexagonal translation lattice. In addition to these translations, $\textnormal{p}_3$ contains rotations of order three (rotation angles $2\pi/3$ and $4\pi/3$), while $\textnormal{p}_6$ contains rotations of order six (rotation angles $\pi/3$, $2\pi/3$, $\pi$, $4\pi/3$, and $5\pi/3$). Both fundamental domains $F_{\textnormal{p}_3 \backslash \textnormal{SE}(2)}$ and $F_{\textnormal{p}_6 \backslash \textnormal{SE}(2)}$ have a regular hexagonal shape in the translational plane, with a height of $2\pi/3$ (from $-\pi/3$ to $\pi/3$) and $\pi/3$ (from $-\pi/6$ to $\pi/6$), respectively. 

\begin{figure}[t]
\centering
\includegraphics[width=0.4\textwidth]{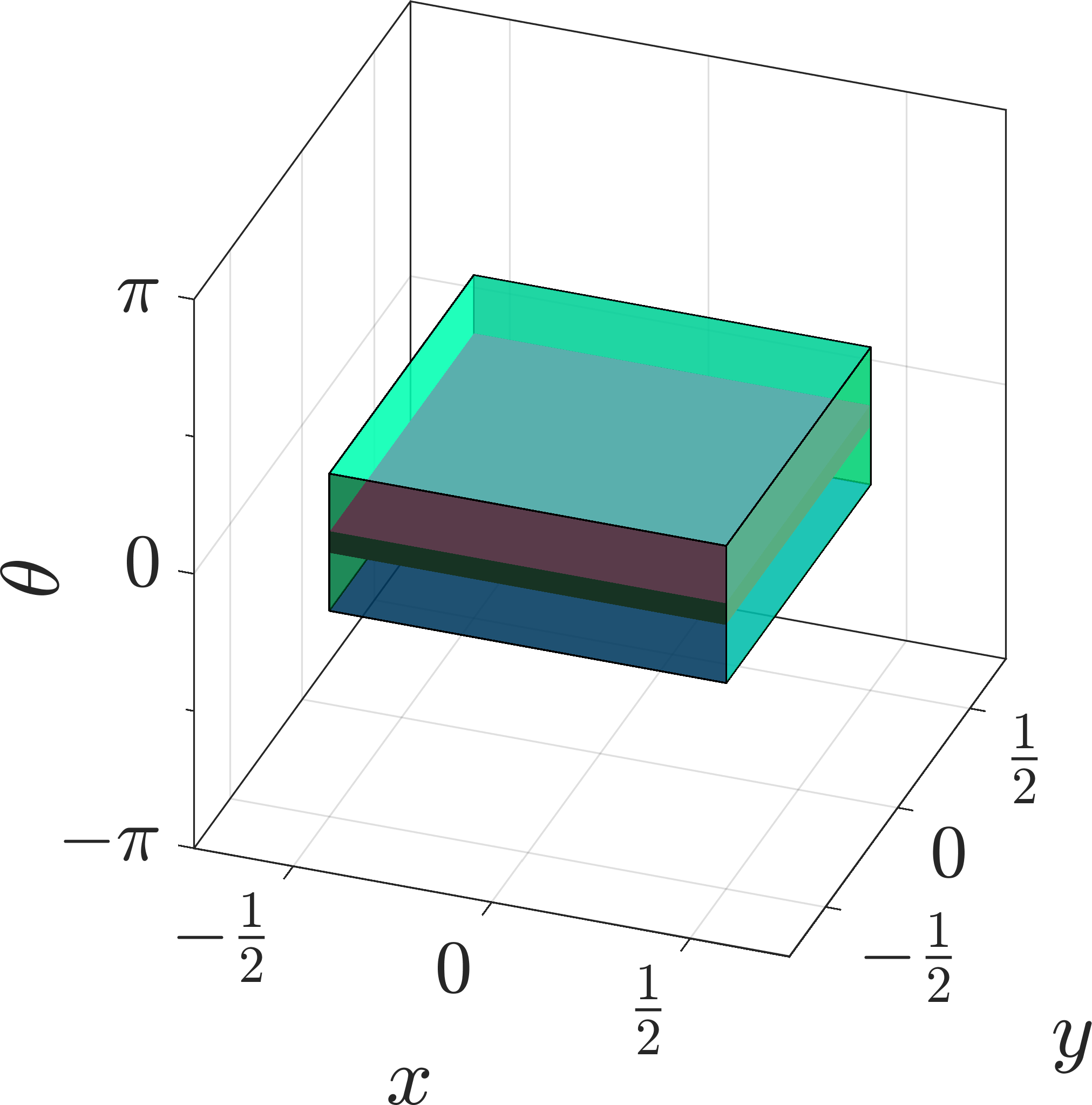}
\caption{Center Voronoi cell $F_{\Gamma \backslash \textnormal{SE}(2)}$ in single-coset space (emerald region) based on an instance of the wallpaper group $\Gamma = \textnormal{p}_4$, and center Voronoi cell $F_{\Gamma \backslash \textnormal{SE}(2) / \Delta}$ in double-coset space (ruby region) with $\Delta = \textnormal{C}_5 \ltimes \{\bm{0}\}$ as the group of rotations of order five.}
\label{fig:p4_shard}
\end{figure}

As an example of a planar-motion alphabet similar to the one described in Section \ref{sec:motion_alphabet}, let us consider the double-coset space $\Gamma \backslash \textnormal{SE}(2) / \Delta$ with $\Gamma \coloneqq \textnormal{p}_4$ and $\Delta \coloneqq \textnormal{C}_{2n-1} \ltimes \{\bm{0}\} < \textnormal{SE}(2)$, where
\begin{equation}\label{eq:roots}
\textnormal{C}_{2n-1} \,\coloneqq\, \left\{
\begin{bmatrix}
\begin{array}{cc}
\cos\theta_j & -\sin\theta_j \\
\sin\theta_j & \cos\theta_j
\end{array}
\end{bmatrix} \,:\, \theta_j = \frac{2 \pi j}{2n-1},~ j = 0,\dots,2(n-1)
\right\} \,<\, \textnormal{SO}(2)
\end{equation}
is the group of rotations of order $2n-1$ 
($n \in \mathbb{N}$).
It is $\Gamma \cap \Delta = \{e\}$ and so we can construct a fundamental domain $F_{\Gamma \backslash \textnormal{SE}(2) / \Delta}$ as a Voronoi-like cell using \eqref{eq:pre_Voronoi_double_coset}. This is shown in Figure \ref{fig:p4_shard} for $n = 3$. In fact, because the left-invariant metric used here is also invariant under purely rotational actions from the right, the fundamental domain $F_{\Gamma \backslash \textnormal{SE}(2) / \Delta}$ is a classical Voronoi cell here, too (again as in the case of $\textnormal{SO}(3)$, cf.\ \eqref{eq:SO(3)_double_coset_Voronoi}). Analogously as in Section \ref{sec:motion_alphabet}, we can use the alphabet $\Gamma \times \Delta < \textnormal{SE}(2)^2$ for an equivolumetric quantization of $\textnormal{SE}(2)$, which at the same time allows for a very uniform sampling of the group. Because the scaling of the 
translational lattice in the wallpaper groups is arbitrary, we can make the alphabet $\Gamma \times \Delta$ arbitrarily fine by reducing the translational scaling in $\textnormal{p}_4$ and increasing the parameter $n$ above.

To illustrate the use of the planar-motion alphabets constructed above, let us consider the $\textnormal{SE}(2)$ trajectory $g(\tau) \coloneqq (R(\tau),\bm{t}(\tau))$, $\tau \in [0,2\pi)$, where $R(\tau)$ is a rotation by an angle of $\tau$ and
\begin{equation*}
\bm{t}(\tau) \,\coloneqq\, \Big[4\cos\tau,\, 6\Big(\frac{\tau}{2\pi}\Big) - 3\Big]^\textnormal{T}.
\end{equation*}
We may discretize $g$ at the five equidistant time points $\pi (1 + 2k / 5)$, $k = -2,\dots,2$. As a motion alphabet for $\textnormal{SE}(2)$, let us use $\Gamma \times \Delta = \textnormal{p}_4 \times \textnormal{C}_5$. We can denote the elements of $\textnormal{C}_5$ by $\delta_j$ with the index $j$ as in \eqref{eq:roots}. Let us denote the elements of $\textnormal{p}_4$ as $\gamma_{lmn}$, where $m \in \mathbb{Z}$ and $n \in \mathbb{Z}$ denote the translation in $x$ and $y$ direction, respectively, while $l \in \{0, 1, 2, 3\}$ indicates a rotation by an angle of $l \pi / 2$. The continuous motion trajectory $g$ can now be expressed as the sentence $(\gamma_{2,3,-2},\delta_3)$, $(\gamma_{2,-1,-1},\delta_4)$, $(\gamma_{2,-4,0},\delta_0)$, $(\gamma_{2,-1,1},\delta_1)$, $(\gamma_{2,3,2},\delta_2)$.

\begin{figure}[t]
\centering
\includegraphics[width=0.485\textwidth]{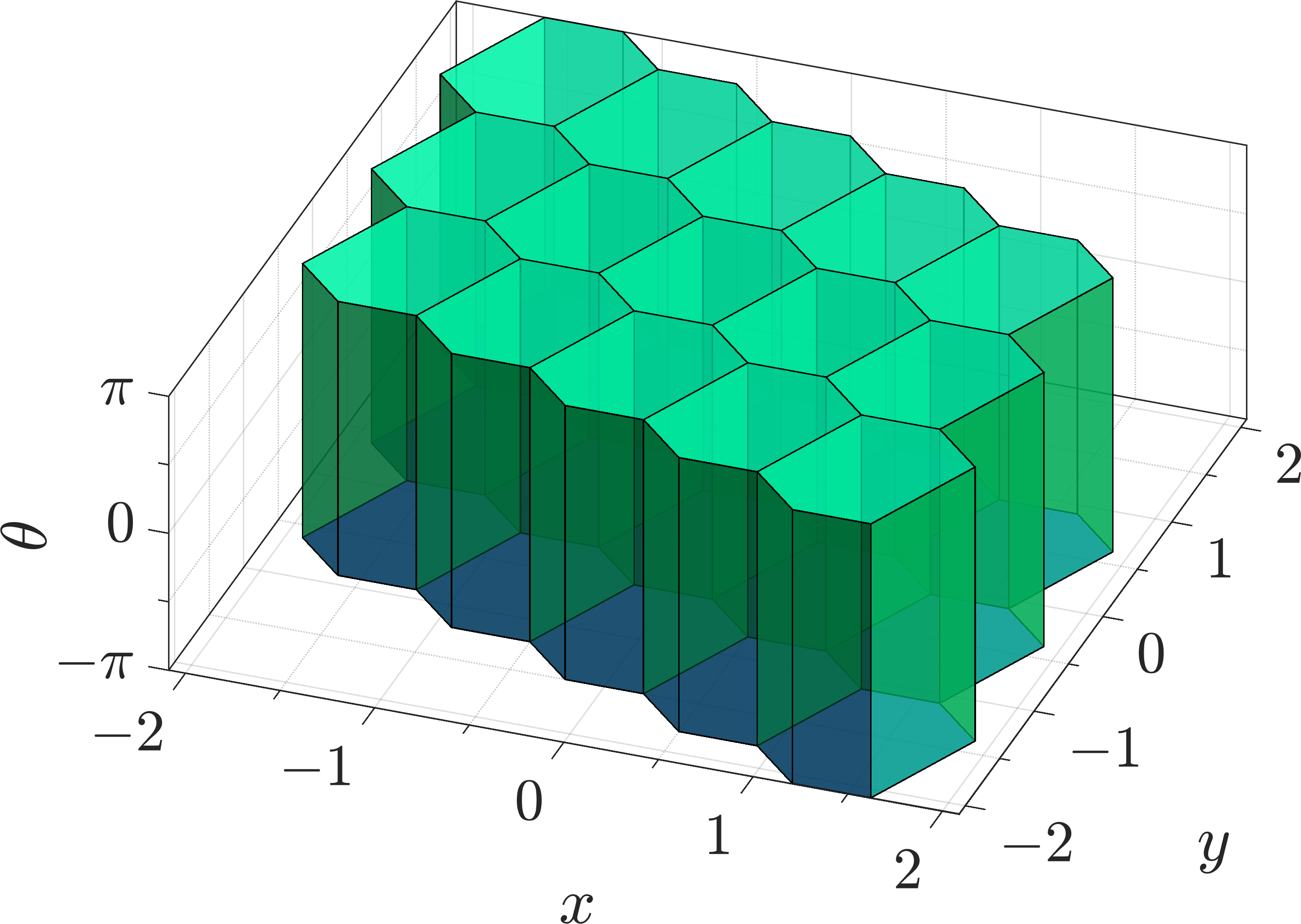}\hfill
\includegraphics[width=0.485\textwidth]{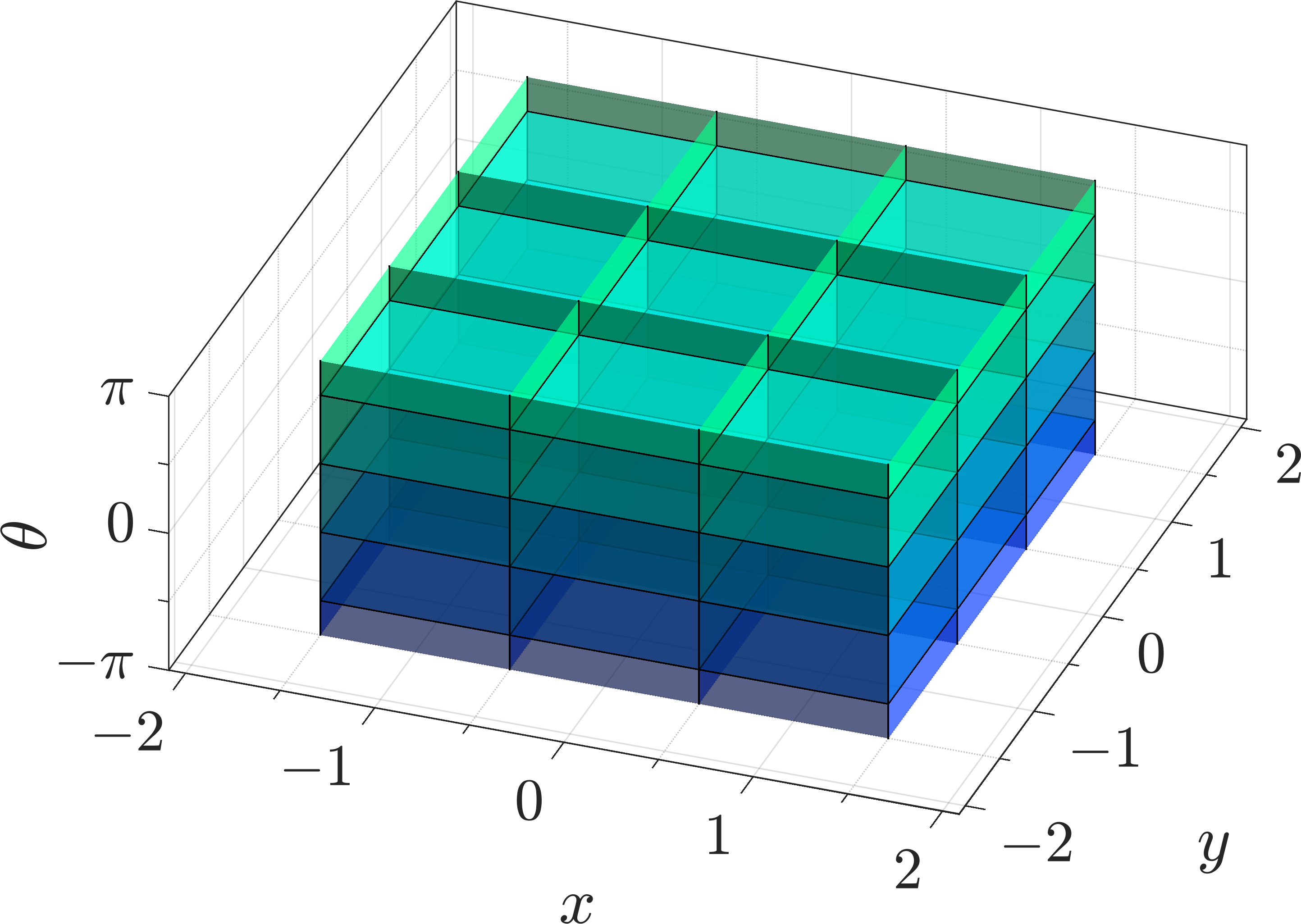}\hfill
\caption{Tesselation of $\textnormal{SE}(2)$ illustrated in $\mathbb{R}^2 \times (-\pi,\pi)$, based on the fundamental domain (left) $F_{\textnormal{p}_1 \backslash \textnormal{SE}(2)}$ and (right) $F_{\textnormal{p}_4 \backslash \textnormal{SE}(2)}$ (conceptual plot).}
\label{fig:SE(2)_tesselation}
\end{figure}

To close this section, we shall discuss how such decoding problem can be solved efficiently. We can use the same coarse-to-fine search scheme that we used in the purely rotational case in the previous section: In a first step, for a given element $g \in \textnormal{SE}(2)$, we find $\gamma \in \textnormal{p}_4$ such that $g \in \gamma F_{\Gamma \backslash \textnormal{SE}(2)}$. This step is particularly easy in the case of $\textnormal{p}_4$ when compared with, \textit{e.\,g.}, the group $\textnormal{p}_1$ (with anisotropic translational lattice). In fact, as implied by Figure \ref{fig:SE(2)_tesselation}, in the case of $\textnormal{p}_4$ the above step can be realized by appropriately rounding off (in the decimal sense) the translational components of $g$, as well as the rotation angle. In the case of $\textnormal{p}_1$, on the other hand, we would have to compute the distances of $g$ to the Voronoi centers. In a second step, we search for the shifted fundamental domain $\gamma' F_{\Gamma \backslash \textnormal{SE}(2) / \Delta} \delta$ containing the pulled-back element $\gamma^{-1} g$ by a purely rotational search. The decomposition of $g$ then reads $(\gamma \gamma') Q \delta$ with $Q = (\gamma \gamma')^{-1} g \delta^{-1} \in F_{\Gamma \backslash \textnormal{SE}(2) / \Delta}$. Of coarse it is also possible to first treat the translational part of $g$, and then the rotational part, say.
With appropriate modifications, the above coarse-to-fine decoding scheme can also be used with the fine alphabet 
for $\textnormal{SE}(3)$ developed in 
Section \ref{sec:motion_alphabet}. 

\section{Conclusion}
\label{sec:conclusion}
Robot tasks involve continuous motions in space. The quantization of these motions by introducing a class of motion alphabets has been established in this paper. With such an alphabet, continuous motion trajectories can be captured with finite words/sentences. It was demonstrated in some examples how the possibility to construct fundamental domains for coset and double-coset spaces as Voronoi or Voronoi-like cells can be used to solve this 
decoding or ``signals to symbols'' problem efficiently via a 
coarse-to-fine search scheme. The alphabets developed here will be used in the future to facilitate the connection between advances in artificial intelligence (such as the use of artificial neural networks) and physical robots acting in the world.

\section*{Conflicts of interest}
The authors declare that there is no conflict of interest regarding the publication of this article.

\section*{Funding statement}
This work was performed under Office of Naval Research Award N00014-17-1-2142 and National Science Foundation grant CCF-1640970. The authors gratefully acknowledge the supporting agencies. The findings and opinions expressed here are only those of the authors, and not of the funding agencies.

\bibliography{references}
\bibliographystyle{my_apa}

\end{document}